\documentclass[journal]{IEEEtran}
\usepackage{times}
\usepackage{epsfig}
\usepackage{amsmath}
\usepackage{amssymb}
\usepackage{graphicx}
\usepackage{bm}
\usepackage{caption}
\usepackage{float}  %设置图片浮动位置的宏包
\usepackage{subfigure}
\usepackage{graphbox}
\usepackage{ amssymb }
\usepackage{multirow}
\usepackage{hyperref}
\usepackage{color}

\usepackage{authblk}
\usepackage[figuresright]{rotating}
\makeatletter
\def\hlinew#1{%
  \noalign{\ifnum0=`}\fi\hrule \@height #1 \futurelet
   \reserved@a\@xhline}
\makeatother
\IEEEoverridecommandlockouts                              % This command is only needed if 
\title{\LARGE \bf
NeuronsGym: A Hybrid Framework and Benchmark for Robot Navigation with Sim2Real Policy Learning
}

\author{Haoran Li$^{1}$, Guangzheng Hu$^{1}$, Shasha Liu$^{1}$, Mingjun Ma$^{1}$, Yaran Chen$^{1}$ and Dongbin Zhao$^{1}$% <-this % stops a space
\thanks{$^{1}$Haoran Li, Guangzheng Hu, Shasha Liu, Mingjun Ma, Yaran Chen and Dongbin Zhao are with the State Key Laboratory of Multimodal Artificial Intelligence Systems, Institute of Automation, Chinese Academy of Sciences, Beijing, 100190, China, and also with the University of Chinese Academy of
Sciences, Beijing, China (email : lihaoran2015@ia.ac.cn, huguangzheng2019@ia.ac.cn, liushasha2020@ia.ac.cn, mingjun.ma@ia.ac.cn, chenyaran2013@ia.ac.cn, dongbin.zhao@ia.ac.cn)}%
\thanks{This work is supported by the National Natural Science Foundation of China (NSFC) under Grants No. 62103409, No. 62136008 and No. 62293545. }
}

\IEEEoverridecommandlockouts

\IEEEpubid{\begin{minipage}{\textwidth}\ \centering
		Copyright \copyright 2024 IEEE. Personal use of this material is permitted.  Permission from IEEE must be obtained for all other uses, in any current or future media, including reprinting/republishing this material for advertising or promotional purposes, creating new collective works, for resale or redistribution to servers or lists, or reuse of any copyrighted component of this work in other works.
\end{minipage}}

\begin{document}

\maketitle
% \thispagestyle{empty}
% \pagestyle{empty}

%%%%%%%%%%%%%%%%%%%%%%%%%%%%%%%%%%%%%%%%%%%%%%%%%%%%%%%%%%%%%%%%%%%%%%%%%%%%%%%%
\begin{abstract}

   The rise of embodied AI has greatly improved the possibility of general mobile agent systems. 
   At present, many evaluation platforms with rich scenes, high visual fidelity, and various application scenarios have been developed. 
   In this paper, we present a hybrid framework named NeuronsGym that can be used for policy learning of robot tasks, covering a simulation platform for training policy, and a physical system for studying sim2real problems. 
   Unlike most current single-task, slow-moving robotic platforms, 
   our framework provides agile physical robots with a wider range of speeds and can be employed to train robotic navigation policies.
   At the same time, in order to evaluate the safety of robot navigation, 
   we propose a safety-weighted path length (SFPL) to improve the safety evaluation in the current mobile robot navigation. 
   Based on this platform, 
   we build a new benchmark for navigation tasks under this platform by comparing the current mainstream sim2real methods, 
   and hold the 2022 IEEE Conference on Games (CoG) RoboMaster sim2real challenge\footnote{\url{https://ieee-cog.org/2022/cog_sim2real/index.html}}.  
   We release the codes of this framework\footnote{\url{https://github.com/DRL-CASIA/NeuronsGym}} and hope that this platform can promote the development of more flexible and agile general mobile agent algorithms.

\end{abstract}


%%%%%%%%%%%%%%%%%%%%%%%%%%%%%%%%%%%%%%%%%%%%%%%%%%%%%%%%%%%%%%%%%%%%%%%%%%%%%%%%
\section{INTRODUCTION}

\IEEEPARstart{N}{avigation} and decision-making are basic abilities of general mobile intelligent systems in the physical world. 
Although this field has a long history of research, classic navigation algorithms and SLAM-based decision methods are still vulnerable to perception noise and environment changes. 
In recent years, with the rise of deep reinforcement learning\cite{dqn2015Nature}\cite{hu2024fm3q}\cite{kun2019starcraft}, and embodied AI\cite{Anderson2018OnEO}\cite{Savva2019HabitatAP}\cite{10687514}, 
learning-based navigation and decision-making algorithms are more robust in complex environments\cite{Chen2019LearningEP}\cite{Wijmans2020DecentralizedDP}\cite{Li2020DRLExplore}. 
Since these methods usually require a large amount of trial-and-error data, 
a high-efficiency simulator is necessary for the training of the algorithm due to the safety risk and execution cost of the physical system.
% Due to the limitations of physical system execution security and execution cost, 
% the training process is usually completed on the simulation platform.

In the past few years, a lot of work {has been done to build} a more visually realistic environment and a more flexible robot model, and many benchmarks and platforms for evaluating algorithms have also been developed. For navigation tasks, 
there are some simulation platforms such as MINOS\cite{Savva2017MINOSMI}, Habitat\cite{Savva2019HabitatAP}, Gibson Env\cite{Xia2018GibsonER}, AI2-THOR\cite{Kolve2017AI2THORAI}, 
and hybrid {frameworks} combining {simulators} and physical {robots} such as RoboTHOR\cite{Deitke2020RoboTHORAO}, Duckietown\cite{Paull2017DuckietownAO}, DeepRacer\cite{Balaji2020DeepRacerAR}\cite{10191291}, and our previous work\cite{Ma2023ComparisonOD}. 
There are also some platforms for other tasks, such as Robosuite\cite{Zhu2020robosuiteAM} for robot manipulation, 
VSSS-RL platform\cite{Bassani2020AFF} for robot soccer, etc. 
% Currently, most platforms refer to a single specific task platform which limits the expansion ability of agents to some extent,
% while general mobile intelligent systems usually need to complete different tasks on the same physical platform.  
% % and a single specific task platform limits the expansion ability of agents to some extent; 
% % On the other hand, 
% Moreover, 
Since the popular mobile robots use differential wheels{,} which limit the maximum speed to $0.5\ m/s$\cite{Shah2022GNMAG}, it is difficult to fully study the sim2real gap caused by dynamic differences with the slow robots. 
% In addition, as one of the most popular robot competitions, RoboMaster competitions\cite{robomaster2022} lack a suitable simulation environment for training intelligent algorithms, 
% so the current competitive strategy is still limited to classic robotics.
Therefore, we need a more flexible and agile platform to expand the learning and transferring capabilities of agents.

Reasonable evaluation is the key to improving the {effectiveness} of agents. 
% In the field of mobile robot navigation, the main evaluation metrics include Success Rate (SR) \cite{Shah2022GNMAG}\cite{Shah2020ViNGLO} and Success weighted by (normalized inverse) Path Length (SPL)\cite{Anderson2018OnEO}\cite{Deitke2020RoboTHORAO}.
Success Rate (SR)\cite{Shah2022GNMAG}\cite{Shah2020ViNGLO} and Success weighted by (normalized inverse) Path Length (SPL)\cite{Anderson2018OnEO}\cite{Deitke2020RoboTHORAO} are popular metrics for evaluating learning-based navigation policies. 
% These indicators can effectively evaluate the mobile efficiency of the robot but do not evaluate the safety during navigation. 
Although these metrics can effectively evaluate the navigation efficiency of the robot, they ignore safety during the navigation process.
Especially for the physical system, safety is more important than navigation efficiency. 
The number of collisions is used as the crucial metric to evaluate safety in previous works\cite{Kstner2021ObstacleawareWG}\cite{Truong2022RethinkingSL}. 
However, when a policy has a specified collision probability, 
the count of collisions is larger for the long-path task, and vice versa. 
Therefore, for scenarios with different path lengths, it is difficult to accurately evaluate the navigation safety of the algorithm {using} the number of collisions. 
Moreover, {when continuous collision such as the robot slides against the wall happen,} the number of collisions is difficult to reflect real safety. 
However, this {continuous collision} phenomenon is very common in navigation policies\cite{Kadian2019AreWM}\cite{Batra2020ObjectNavRO}.

\IEEEpubidadjcol

Considering the above problems, we build a mobile robot training and evaluating framework for navigation tasks. 
% Unlike most current platforms, which only contain single-task, the designed framework NeurousGym contains navigation tasks, 
% confrontation tasks and the combination of navigation and confrontation tasks, which can expand the ability of agents to a greater extent. 
We {enrich} the mobile robot navigation performance evaluation system to highlight safer and more efficient robot policies.
Specifically, the contributions of this paper are as follows:
% \begin{itemize}
   1) We build a mobile robot training and evaluating framework for navigation tasks. 
   In the simulation environment of this framework, we build detailed friction force models to provide more accurate dynamics simulation 
   and multiple sensor models to better simulate real data for supporting {diverse} policies. In the physical system of this framework, 
   we build a similar platform and tasks as the simulation, 
   which can directly deploy and verify the {policies} trained in the simulation environment.
   2) We propose a new metric - SFPL, which can be used to evaluate the safety of robot navigation. 
   Compared with the current commonly used number of collisions, it can effectively compare the physical safety of algorithms under various path lengths by weighting collisions and path lengths.
   3) Based on the platform, we compose a new benchmark for navigation tasks and compare the performance of various sim2real algorithms. 
   Moreover, we hold the 2022 IEEE CoG RoboMaster sim2real challenge to promote the development of flexible and agile agents.
% \end{itemize}

\section{Related Work}
% In this paper, we propose a mobile robot training and evaluation platform including navigation and confrontation tasks to study learning-based robot policies and sim2real problems, 
% and propose a metric to evaluate the safety of robot navigation. 
In this section, we review the development of the policy training platforms for navigation and the related platforms used to study the sim2real problems. 
% In addition, we discuss the evaluation metrics of the robot navigation algorithm.

\subsection{Platforms for Navigation Task}
Learning-based mobile robot navigation has a long history of development. 
It gets rid of the dependence on the map by directly establishing the mapping of state to action without relying on 3D perception\cite{10651302}\cite{9892040}. 
In recent years, with the rise of embodied AI, navigation through visual sensors has emerged endlessly\cite{Chen2019LearningEP}\cite{Wijmans2020DecentralizedDP}\cite{Hahn2021NoRN}, 
and corresponding platforms have emerged. For example, Gazebo\cite{Koenig2004DesignAU} has high fidelity for robot simulation, 
but it is not suitable for trial-and-error training of agents due to its low simulation frame rate\cite{Savva2017MINOSMI}. 
ViZDoom\cite{Kempka2016ViZDoomAD}\cite{Shao2018LearningBI} and DeepMind Lab\cite{Beattie2016DeepMindL} also attract some attention in the field of visual navigation due to their excellent simulation efficiency. 
However, its stylized maze environment and weakened robot properties make it not suitable for indoor mobile robot navigation. 
The MINOS\cite{Savva2017MINOSMI}, Habitat\cite{Savva2019HabitatAP}, Gibson Env\cite{Xia2018GibsonER}, AI2-THOR\cite{Kolve2017AI2THORAI} and other environments developed in recent years have high visual fidelity to the environment as well as  
the simulation of the robot. 
{
Similar to these works, the proposed NeuronsGym is designed for robot navigation tasks and also supports visual observation. And like AI2-THOR, we also use Unity as the physics engine. Unlike these simulators, we focus more on robot dynamics and sensor data distribution simulation compared to visual simulation. In addition, we highlight the safety in navigation during the evaluation, which other simulators do not possess. More importantly, we emphasize that the proposed framework is a hybrid architecture that covers both simulation training and physical deployment. This is not available in the aforementioned simulators.}

% The above platforms are only limited to research in the virtual environment 
% and do not configure a physical robot to study the transfer from the simulation to the real {world}. 
RoboTHOR\cite{Deitke2020RoboTHORAO} configures a LoCoBot\cite{Murali2019PyRobotAO} robot on the basis of AI2-THOR to facilitate the transfer of visual perception and navigation algorithms trained in the simulation environment to the real robot tasks. 
Sim2Real Challenge\cite{igibsonchallenge2020} held {at} the 2020 CVPR uses {the} iGibson simulator and LoCoBot robot to study how visual navigation algorithms can be more efficiently transferred from the simulation to the real environment. 
In addition, DeepRacer\cite{Balaji2020DeepRacerAR} and Duckietown\cite{Paull2017DuckietownAO} provide an open-source low-cost standardized hardware platform. 
Combining with the tracking navigation task, 
they are used to study the impact of the transfer of visual perception and dynamics from the simulation to the real environment. 
Similar to these works, our platform also provides a visual sensor and dynamic parameter interface for studying reality gaps through joint perception and dynamics. 
Compared with the maximum moving speed limit of $0.5 m/s$ for most indoor mobile robots\cite{Murali2019PyRobotAO}\cite{Shah2022GNMAG}, 
we provide a wider speed interval to study the sim2real gap caused by different execution speeds.

Metrics are crucial for evaluating {the} algorithm's performance.
In the current works\cite{Savva2017MINOSMI}\cite{Shah2022GNMAG} for studying learning-based navigation algorithms, 
% the ratio of the robot successfully reaching the goal to the total number of trials, also named SR,
SR 
is one of the most commonly used indicators to evaluate the performance of the navigation algorithm. 
% It is the ratio of the robot successfully reaches the goal to the total number of trials. 
There are other similar indicators, such as timeout\cite{Ryu2022ConfidenceBasedRN}, which are used to indirectly evaluate the robot's ability to reach the target. 
Since these indicators are difficult to reflect the navigation process, the navigation efficiency can also be evaluated with such indicators as path length\cite{Deitke2020RoboTHORAO}, 
average speed, and consumed time. Moreover, there are also indicators that combine SR with navigation efficiency, 
such as SPL\cite{Anderson2018OnEO}\cite{Truong2022RethinkingSL} and Success weighted Completion Time (SCT)\cite{Yokoyama2021SuccessWB}. In addition to navigation capability, 
navigation safety is a more important performance in physical systems\cite{9882307}, since it directly affects the health of robot systems, 
and even affects the safety of navigation participants (such as pedestrians). In the navigation system of indoor mobile robots, 
collision times and collision {rates} \cite{Ryu2022ConfidenceBasedRN} are the main indicators used to evaluate the safety of the system. Similar to the SR, 
it can only generally express the navigation safety results but is difficult to describe what happened during the navigation process. 
The number of collisions as an indicator of safety assessment also faces some problems. 
When a policy has a specified collision probability, the count of collisions is larger for the long-path task, and vice versa. 
Therefore, for scenarios with different path lengths, it is difficult to accurately evaluate the navigation safety with the number of collisions. 
Another dilemma is to judge the occurrence of a collision. {A common one-on-one} collision means that a robot contacts or approaches an obstacle and then leaves. 
However, in many robot navigation policies, {the phenomenon of continuous collision such as sliding against obstacles appears, }
which {makes it} difficult to evaluate safety with the number of collisions. 
Therefore, we need a more reasonable navigation safety evaluation metric to build a more secure navigation agent.

\begin{figure*}
   \centering  
   \includegraphics[width=0.9\linewidth]{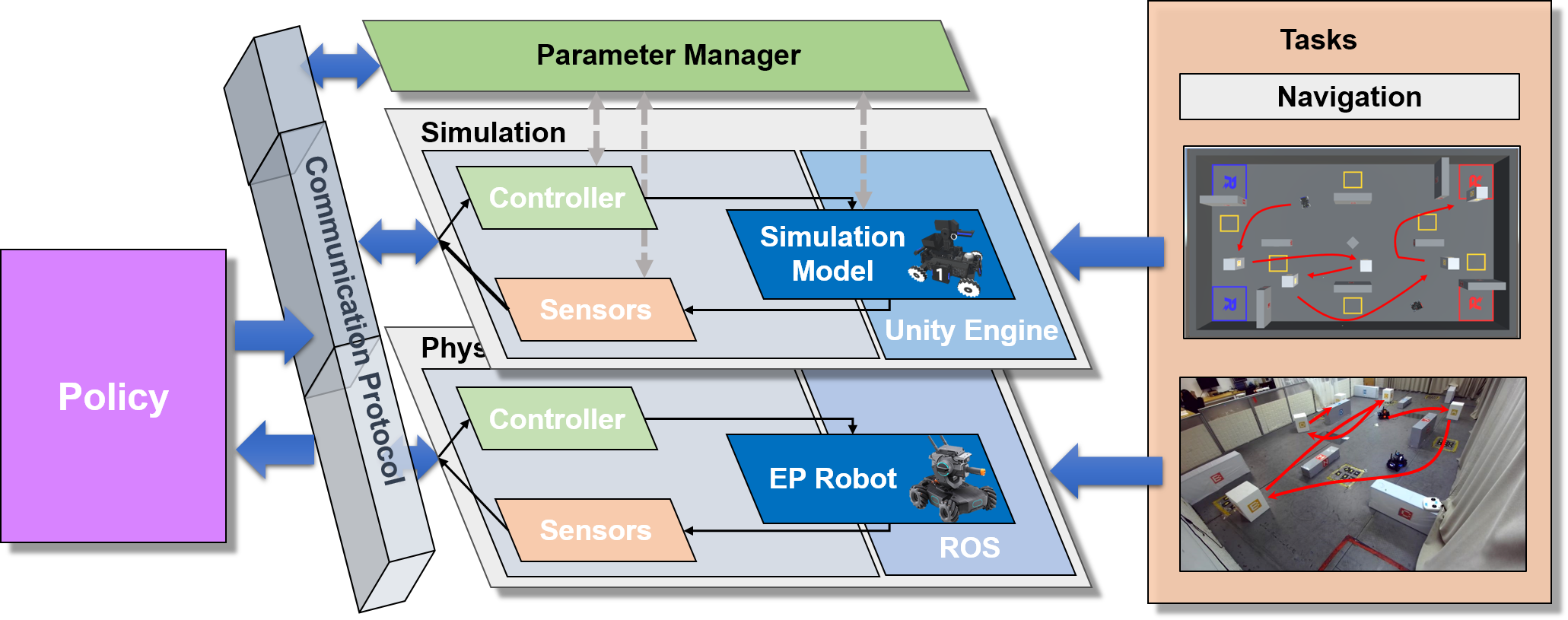}
   \caption{Overview of the hybrid framework - NeuronsGym. 
   The framework is composed of simulation and physical systems. 
   Agents can interact with simulation systems or physical systems through communication protocols to achieve agent training or evaluation. 
   The agent policy can access the parameter manager to adjust the parameters of the robot model or environment in the simulation system. 
   In addition, the same scenario and task are set in each system to study the sim2real of the robot policy.
   }
   \label{fig_framework}
   \vspace{-0.5cm}
\end{figure*}

\subsection{Methods and Platforms for Sim2Real Problems}
Transferring the trained policy {from} the simulation environment to the real environment is a challenging task, 
since the gap between the simulation and the real degenerates the performance.  
There has been much research on the sim2real gap. 
% These works can be divided into two categories: one is for visual texture, 
Some works are mainly to solve the problem of generalization of the models caused by the distribution shift between the virtual data and the real data\cite{Ren2019DomainRF}\cite{Sadeghi2016CAD2RLRS}. 
This kind of method is not the focus of this paper. 
% The other is dynamic model of robots, 
Another work mainly solves the problem of model performance degradation caused by the dynamic difference between the simulation robot and the real robot. 
The common methods for this kind of problem include system identification\cite{Punjani2015DeepLH}\cite{Zhu2017FastMI}, 
domain randomization\cite{Peng2017SimtoRealTO}\cite{Chebotar2018ClosingTS}\cite{Mehta2019ActiveDR}, action transformation\cite{Hanna2017StochasticGA}\cite{Hanna2021GroundedAT}, and so on. 
Due to the lack of a suitable physical platform,
some research work only relies on the sim2sim method of simulation environment to indirectly carry out sim2real problem research\cite{Mehta2020AUG}\cite{Jiang2021SimGANHS}. 
This method simplifies the complexity of the problem while ignoring the influence of uncontrolled factors such as control delay, 
observation, and execution noise after transferring. In addition, there are many works combining the simulation environments with the physical robot. 
For example, grasping operation of {a} robot arm\cite{Deitke2020RoboTHORAO}, manipulation of a dexterous hand\cite{Li2019LearningTS}, motion control of a quadruped robot\cite{Kumar2021RMARM}\cite{Lee2020LearningQL}, etc. 
In addition, many sim2real-related competitions have been held, such as the aforementioned 2020 CVPR sim2real challenge, 
the 2021 NeurIPS AI Driving Olympics\cite{aidriving2021} based on the Duckietown platform, and the 2022 ICRA sim2real challenge\cite{rmus2022} for mobile grasping robots. 
It should be pointed out that although the simulation environment and physical robots are used in the ICRA sim2real challenge, 
the motivation for this challenge is still the classical robot algorithm based on traditional planning and control algorithms. 
Therefore, there is a significant difference between sim2real discussed in this paper{,} which focuses on learning-based algorithms.
{
Our NeuronsGym provide the detailed robot dynamics model and the sensor models, and supply the interface for the sim2real methods. Therefore, different parameter distributions can be set through domain randomization, domain adaptation, or other sim2real methods to obtain policies with different generalization performance.
}

\section{NeuronsGym}
% Unlike most current platforms, which only contain a single task, 
We present a hybrid framework named NeuronsGym, designed to train learning-based robot policies  
and evaluate the transfer ability and generalization of algorithms. 
In this framework, we provide a simulator containing a robot and an arena, as well as a corresponding physical system.
We design robot navigation as well as corresponding metrics for training and evaluating robot policies. 
In the simulation environment, we have configured a variety of sensors and rich interfaces {for} robot dynamics to support the research on sim2real issues. 
The overview of this framework is shown in Fig. \ref{fig_framework}.
% Next, we will introduce the framework in detail.

\subsection{Tasks}
As shown in Fig. \ref{fig_framework}, we {design} a robot navigation task in NeuronsGym. 
% As the user of the environment, you can use a robot navigation task, or robot confrontation task alone, and also combination of the two tasks.
% \subsubsection{Robot Navigation Task}
Similar to the PointGoal\cite{Anderson2018OnEO} navigation task, 
the robot starts from the randomly generated starting pose and navigates to the specified goal only according to its sensor data. 
It should be noted that in our task setting, the position of the goal is also randomly generated. 
{
Different from the PointGoal task, 
the task of navigating to the goal can only be completed with a certain distance and angle, which is named "activation". Specifically, when the distance from the goal to the robot is less than 1.0 $m$, and the angle between the robot and the goal is less than 30 degrees, the robot activate the goal.}
In addition, the agent needs to activate 5 goals in alphabetical order to complete the navigation task. It should be noted that our goals are not virtual points but obstacle blocks with collision attributes. Therefore, in each navigation trial, the layout of the arena is different due to the random placement of the goal blocks.

\subsection{Simulation Platform}
In the NeuronsGym framework, we build a simulation platform based on Unity3D to train agent policies. This simulation platform includes the arena required for different tasks, 
the simulation robot models, the controllers, and a variety of sensors.

\subsubsection{Arena}
% The layout of the arena in the simulator is borrowed from the RMUA. 
{
The arena is a $5 m \times 8 m$ rectangular area with several obstacles. These obstacls have different heights, 10 $cm$ and 20 $cm$ respectively. 
The 20 $cm$ height obstacles can block both the camera field of view and the LiDAR field of view, while the 10 $cm$ height obstacle can only block the LiDAR field of view. 
Around the obstacles, we arrange some visual tags to provide richer visual features to the robot. 
In addition, we place 5 blocks with a height of 20 $cm$ and a width of 10 $cm$ in the field as the goals to be activated. Around the blocks, we paste 5 letters from A to E.
The difference between obstacles and blocks is that the position of obstacles is fixed and unchanging, while the position of blocks varies in each trail and may be changed by robots.
}

\subsubsection{Robot}
The simulation robot is based on the infantry of RoboMaster EP.
% The overall architecture of the simulation robot is shown in Fig. \ref{fig_robot}.
Since the RoboMaster EP is a commercial product and its mechanical drawings are not open source, we model the core mechanism of the robot, 
including the chassis part with the Mecanum wheel, the two degrees of freedom gimbal part, and the firing unit. For the moving simulation of the chassis, 
considering the simulation efficiency, we abandon the fully dynamical simulation of driving the robot motion with the friction between the rollers of the wheel and the ground in the simulator, 
but directly set the {robot's} linear and angular velocities. In order to ensure the consistency of the motion characteristics between the simulator and the physical robot, 
we build a mathematical model for the simplified part of the speed calculation, so as to realize the influence of many factors on the speed, 
such as friction parameters, motor characteristics, and control parameters.

\begin{figure}
   \hspace{0.5cm}
	\centering 
   \includegraphics[width=0.45\textwidth]{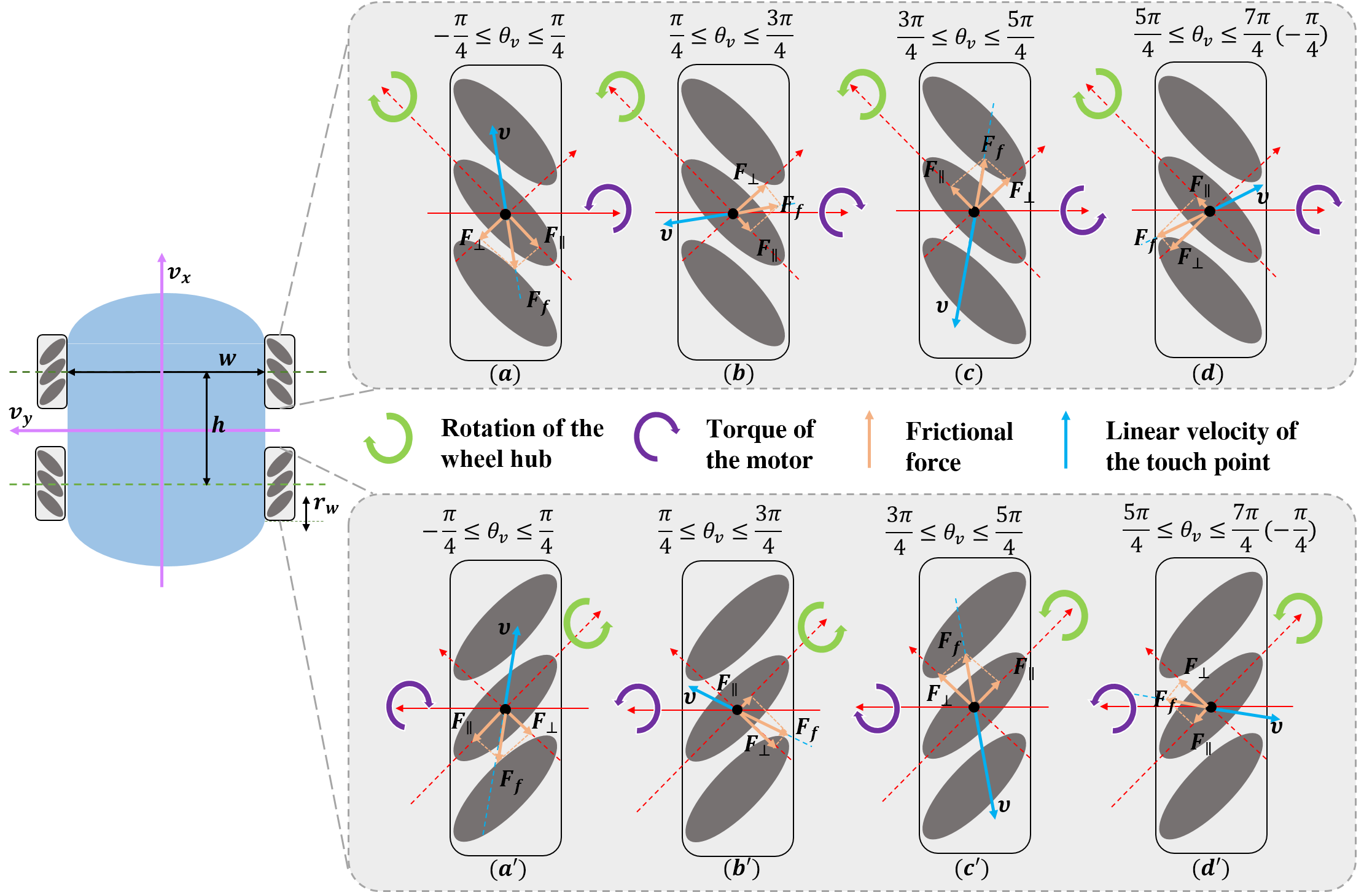} 
   
	\caption{Friction force analysis of the robot wheel.}
   \label{fig_friction}
   \vspace{-0.5cm}
\end{figure}

% Based on the characteristic curve of the motor M3508 used in the physical robot, the torque generated under the current action can be calculated as
Specifically, for any of the wheels $i$, when the current inputting to the drive motor is $I_i(t)$, 
the wheel rotational speed $\omega_i (t+1)$ at the next time can be calculated according to the current wheel rotational speed $\omega_i (t)$
% where $C_T$ is the motor characteristic parameter. 
% According to the angular momentum theorem, the acceleration $\dot{\omega}_i$ of the wheel rotation can be obtained under the action of the torque provided by this motor.
\begin{equation}
    \omega_i (t+1) = \omega_i (t) + \int_t^{t+1} \frac{C_T I_i(\tau) - 4 F_{f_i}(\tau) r_w / M}{\rho_w} d \tau.
\end{equation}
Here $r_w$ and $\rho_w$ are the radius and the rotational inertia of the wheel, respectively. 
$C_T$ is the motor characteristic parameter. $M$ is the mass of the robot.
Here $F_{f_i}$ is the friction between the wheel $i$ and the ground. 
It should be noted that when the wheel changes from the motionless to the rotating state, 
the friction will change from static friction to dynamic friction. The process is calculated as follows
\begin{equation}
   F_{f_i}(\tau)  = \begin{cases}
      C_T I_i(\tau), & | \omega_i | \leq \omega_e \\
      f_{d_i}. & | \omega_i | > \omega_e\\
   \end{cases}
\end{equation}
Here $\omega_e$ is the threshold. 
$f_{d_i}$ is the dynamic friction parameter. It includes the magnitude and direction of dynamic friction, 
which is composed of sliding friction $f_{\shortparallel}$ and rolling friction $f_{\bot}$ of the roller. When the speed of the robot changes, 
the direction of sliding friction and rolling friction will also change. 
The robot is composed of two different groups of wheels, 
and the friction direction between the groups is different. 
Specifically, when the velocity direction of the robot is known as $\theta_v$, we can analyze the friction of each wheel, as shown in Fig. \ref{fig_friction}. 
For the right front wheel $i=1$ and the left rear wheel $i=3$, the friction can be calculated by the following
\begin{equation}
   f_{d_i} = \begin{cases}
      -f_{\bot} \sin (\theta_v - \frac{\pi}{4}) + f_{\shortparallel} \cos (\theta_v - \frac{\pi}{4}), & -\frac{\pi}{4} \leq \theta_v \leq \frac{\pi}{4} \\
      f_{\bot} \sin (\theta_v - \frac{\pi}{4}) + f_{\shortparallel} \cos (\theta_v - \frac{\pi}{4}), & \frac{\pi}{4} \leq \theta_v \leq \frac{3\pi}{4} \\
      f_{\bot} \sin (\theta_v - \frac{\pi}{4}) - f_{\shortparallel} \cos (\theta_v - \frac{\pi}{4}), & \frac{3\pi}{4} \leq \theta_v \leq \frac{5\pi}{4} \\
      -f_{\bot} \sin (\theta_v - \frac{\pi}{4}) - f_{\shortparallel} \cos (\theta_v - \frac{\pi}{4}). & \frac{5\pi}{4} \leq \theta_v \leq \frac{7\pi}{4} \\
   \end{cases}
\end{equation}
For the left front wheel $i=2$ and the right rear wheel $i=4$, 
the calculation process is slightly different since the arrangement direction of their roller is different from the previous wheels $i=1,3$. 
Specifically, it can be calculated by the following 
\begin{equation}
   f_{d_i} = \begin{cases}
      f_{\bot} \cos (\theta_v - \frac{\pi}{4}) - f_{\shortparallel} \sin (\theta_v - \frac{\pi}{4}), & -\frac{\pi}{4} \leq \theta_v \leq \frac{\pi}{4} \\
      f_{\bot} \cos (\theta_v - \frac{\pi}{4}) + f_{\shortparallel} \sin (\theta_v - \frac{\pi}{4}), & \frac{\pi}{4} \leq \theta_v \leq \frac{3\pi}{4} \\
      -f_{\bot} \cos (\theta_v - \frac{\pi}{4}) + f_{\shortparallel} \sin (\theta_v - \frac{\pi}{4}), & \frac{3\pi}{4} \leq \theta_v \leq \frac{5\pi}{4} \\
      -f_{\bot} \cos (\theta_v - \frac{\pi}{4}) - f_{\shortparallel} \sin (\theta_v - \frac{\pi}{4}). & \frac{5\pi}{4} \leq \theta_v \leq \frac{7\pi}{4} \\
   \end{cases}
\end{equation}
Fig. \ref{fig_friction} shows the detailed force analysis of two groups of wheels under four conditions. 
In the above calculation, we use the velocity direction of the robot. 
This direction is the linear velocity direction of the contact point between the wheel and the ground. 
It is the result of the superposition of the wheel's angular velocity and the robot's angular velocity. 
The actual moving speed direction of each wheel can be calculated {using} the triangle cosine theorem. 
The calculation process is not complicated, so we will not introduce it in detail here.

It should be noted that in the above modeling process, although we have simplified the process for some special cases, 
such as the wheel slippage phenomenon and the linear approximation of the motor characteristic curve, 
our experimental results show that the above model can perform a better simulation for the non-limit motion cases.

\subsubsection{Sensors}
% On the physical robot, we installed LIDAR, odometer and camera sensors. Therefore, in the simulator, we have also constructed simulation models of these sensors.
In the simulator, we mainly provide three kinds of sensors: LiDAR, odometer, and camera. 
In addition, the speed, pose, collision, and other information of the robot can be directly obtained through Unity3D.

\textbf{LiDAR }
We build a virtual LiDAR with a single laser. The scanning angle range of this virtual LiDAR is $[-135, 135]$, 
and the angular resolution is 4.5 degrees. Therefore, each frame contains 61 scanning points. 
In the simulator, we can obtain the precise distance from the LiDAR installation position to the obstacle. 
However, real LiDAR measurements usually contain outliers and measurement noise. 
Therefore, we use Gaussian distribution to build the LiDAR noise model. 
For any distance measurement noise $n^L_i$, its generation probability $p(n^L_i)$ is
\begin{equation}
   p(n^L_i) = \frac{1}{\sqrt{2 \pi} \sigma} \exp (\frac{(n^L_i - \mu_l)^2}{2 \sigma_l^2}).
\end{equation}
Here $\mu_l$ and $\sigma_l$ are the mean and standard deviation of the Gaussian distribution, respectively.
In addition to normal data, anomalous data can be caused by the failure to receive reflected data due to the surrounding environment material, 
certain special incidence angles between the LiDAR and the object, and so on. Considering the randomness of anomalous data, 
we use Poisson distribution to build the LiDAR anomaly model. 
For any frame of LiDAR data, the probability of $k$ abnormal points is
\begin{equation}
   p(k) =  \frac{\lambda^k}{k!} \exp (-\lambda), k = 0, 1, \cdots
\end{equation}
where $\lambda$ is the Poisson distribution parameter. 
After the number of abnormal data is determined by sampling, we randomly select data from the point cloud and process them as abnormal values.

\textbf{Odometer }
As a common position measurement sensor on most mobile robots, encoders are commonly used to generate odometer data. 
In the previous robot motion modeling, we maintain the rotational speed of each wheel. 
Through the Mecanum wheel kinematics model, we can get the velocity $(\tilde{v}_x(t), \tilde{v}_y(t), \tilde{v}_w(t))$ of the robot according to the wheel speeds.
\begin{align}
    \tilde{v}_x(t) = \frac{(\tilde{\omega}_3(t) + \tilde{\omega}_4(t)) r_w}{2} \nonumber \\
    \tilde{v}_y(t) = \frac{(\tilde{\omega}_3(t) - \tilde{\omega}_1(t)) r_w}{2} \label{eq_vel} \\
    \tilde{v}_w(t) = \frac{(\tilde{\omega}_2(t) - \tilde{\omega}_3(t)) r_w}{(h+w)} \nonumber
\end{align}
In the above equation, $\tilde{\omega}_i(t)$ is the observation of the wheel rotational speed. 
$h$ and $w$ are the wheelbase and wheel tread of the robot, respectively.
Here we simulate the measurement process of the encoder with Gaussian noise.
% It is assumed that the wheel encoder measurement error on each wheel conforms to Gaussian noise. Then, 
The measurement is
\begin{equation}
   \tilde{\omega}_i(t) = \omega_i(t) + n^e, n^e \sim \mathcal{N}(\mu_e, \sigma_e)
\end{equation}
where $\mu_e$ and $\sigma_e$ are the mean and standard deviation of the measurement noise, respectively. 
In formula \ref{eq_vel}, the velocity of the robot speed is measured by the wheel encoders. 
By integrating the velocity, we get the simulated odometer measurement of the robot.
\begin{align}
   x(t) &= \int_0^t \cos(\theta(\tau)) \tilde{v}_x(\tau) - \sin(\theta(\tau)) \tilde{v}_y(\tau) d \tau \nonumber \\
   y(t) &= \int_0^t \sin(\theta(\tau)) \tilde{v}_x(\tau) + \cos(\theta(\tau)) \tilde{v}_y(\tau) d \tau \nonumber \\
   \theta(t) &= \int_0^t \tilde{v}_w(t) d \tau \nonumber
\end{align}
% It should be noted that the velocity of the robot is in the robot coordinate. 
% Before integral calculation, it is necessary to transfer the velocity to the odometer coordinate.
\begin{figure}
    \hspace{0.5cm}
	\centering  
    \includegraphics[width=0.49\textwidth]{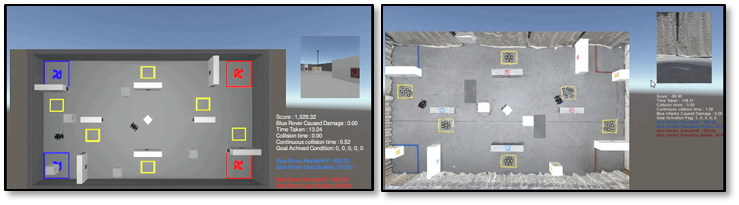} 
   
	\caption{{Simulation arenas. The left is the arena established with the built-in geometric modules in Unity3D, and the right is the arena built by RealtyCapture with the real arena images.}}
    \label{fig_arena}
    \vspace{-0.5cm}
\end{figure}
\textbf{Camera }
Images are becoming core state input in more and more methods, and our proposed simulator also sets the camera sensor to support image output. 
In this section, we mainly rely on the camera sensor that comes with the Unity3D engine. By setting the field of view and focal length, 
we adjust the field of view to be as similar as possible to the camera carried by the physical robot. In terms of image rendering similarity, 
we provide two high-fidelity field models in the arena section to compromise the simulator rendering efficiency and image reproduction. 
{
Considering that the rendering efficiency of the model can greatly affect the simulation efficiency, two arena models are constructed. 
One is a simple model based on Unity3D built-in geometry units, as shown in Fig. \ref{fig_arena}. 
% During the construction of this model, 
% we replicate the physical arena in the simulator based on the layout, the size, and the obstacles with the same color and material. 
Since only the base models built in Unity3D are used, collision detection and image rendering are highly efficient in this platform. 
In this arena model, we do not model the background outside the arena, and the images collected under this model are different from the images collected by the physical robot. 
Another arena model is constructed based on 3D reconstruction software. We pre-capture a large number of high-quality images in the physical arena, 
import them into RealityCapture\cite{realitycapture}, and perform visual 3D reconstruction of the captured arena by matching feature points in the images. 
In the 3D reconstruction process, a large number of detailed grids are used to model the environment in order to restore the physical arena more realistically, 
which can lead to relatively low efficiency of Unity3D in the collision detection and rendering process. 
Users can select the arena model according to the efficiency and characteristics of the designed algorithm.
}

\subsubsection{Controller}
% According to the kinematic model of Mecanum wheel, when the expected velocity at the next moment is $\hat{v}(t+1)=[\hat{v}_x(t+1), \hat{v}_y(t+1), \hat{v}_w(t+1)]$, 
% the desired linear velocity $\hat{v}_{w_i}(t+1)$ of the 4 wheels at the point of contact with the ground can be calculated by the following equation
The main work of the controller is to calculate the current required by the motor according to the input control. 
In our system, the control related to the robot's movement is the expected velocity of the robot. 
In the simulator, we configure a motor for each wheel. 
The current input required by the motor is calculated according to the expected rotational speed of each wheel and the actual rotational speed. 
We use a {Proportional-Integral-Derivative} (PID) controller to calculate the required current. 
First, we need to calculate the expected rotational speed $\hat{\omega}_i(t)$ according to the kinematic model of the robot and the control input $(u_x(t), u_y(t), u_w(t))$ at the current time.
Then PID is used to calculate the current $I_i(t)$ corresponding to each wheel.
\begin{align}
    I_i(t) = k_p (\hat{\omega}_{i}(t)- \omega_{i}(t)) + k_i \int_0^{t}(\hat{\omega}_{i}(\tau)- \omega_{i}(\tau))d \tau \nonumber \\
     + k_d \frac{d(\hat{\omega}_{i}(\tau)- \omega_{i}(\tau))}{d \tau} \nonumber
\end{align}
where $k_p$, $k_i$, and $k_d$ are the proportional, integral, and differential parameters, respectively.

\subsection{Physical Platform}
In the real environment, we have built an arena and a robot system that are the same as the simulation environment. 
Next, we will introduce these two parts respectively.

\subsubsection{Arena}
The arena in the real environment is built in equal proportion to the arena in the simulator. 
The layout of the obstacles and the visual tags are consistent with the simulation environment. 
Different from the simulation environment, the goal blocks in the real environment can be pushed by the robot, 
while the blocks in the simulation environment are fixed.

\subsubsection{Robot and Sensors}
We use the infantry of RoboMaster EP as the robot platform. 
Compared with other commonly used mobile robot platforms (Turtle bot, LoCoBot, etc.), 
RoboMaster EP has a higher speed, bullet shooting function, and bullet sensing device, 
which can better meet the task requirements in the framework. 
In addition to the cameras, odometers, and other sensors already configured on the robot, 
RPLiDAR S2 is installed in front of our robot, and NVIDIA Jetson NX is arranged behind the robot as the computing platform. 
It should be noted that pointcloud distortion of the real LiDAR will happen during the robot's fast movement, 
which does not exist in the simulation environment. Therefore, we correct the pointcloud with an odometer to remove the distortion. 
In addition, compared with the LiDAR in the simulation environment, the angle resolution of RPLiDAR S2 is smaller and the field of view is larger. 
{In order to keep the simulation environment the same, we sample the real pointcloud. In the real robot, the true pose of the robot cannot be directly measured by the sensor.} 
Therefore, an Adaptive Monte Carlo Localization (AMCL) algorithm is used to obtain the true pose of the robot.

\begin{figure*}
   \centering  
   \includegraphics[width=0.95\linewidth]{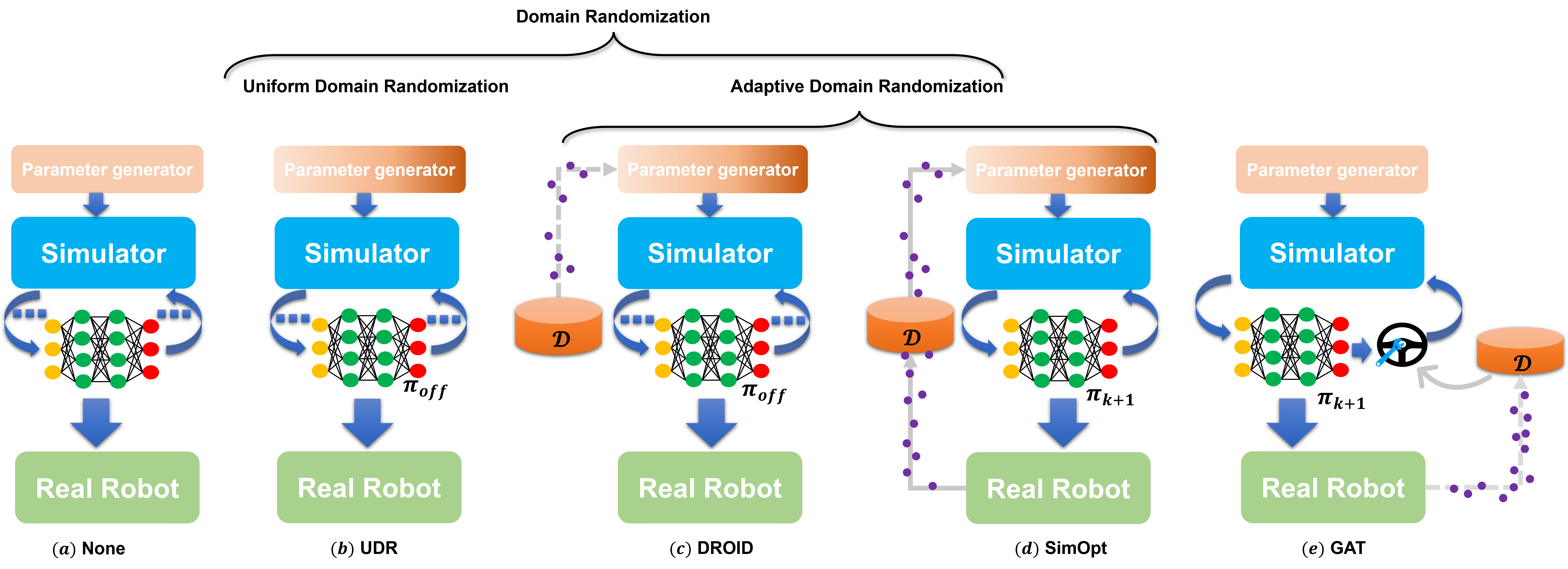}
   \caption{Several different sim2real methods. 
   In the training process, each trial starts by sampling the simulator parameters from the parameter generator, 
   then uses the simulator to train the agent, and then transfers to the real robot. 
   The difference is that the parameter generators of $(a)$ and $(e)$ get the same results each time, and $(b) - (d)$ sample parameters from some distribution, 
   and the results are different each time. Here, $(b)$ does not require actual robot data, $(c)$ requires offline robot data, 
   and $(d)$ requires online robot data. 
   Unlike $(c)$ and $(d)$, $(e)$ uses offline data to learn the action transformer to correct the state generated by the simulator.
   }
   \label{fig_baselines}
   \vspace{-0.5cm}
\end{figure*}

\section{Sim2Real Protocols and Baselines}
Due to the low sampling efficiency and safety of the physical system, most of the current robotic reinforcement learning algorithms are trained in the simulator. 
The policy directly transfers from the simulation environment to the real environment and will face performance degradation {due to} the gap between the simulation environment and the real environment. 
Therefore, how to transfer the trained agents to the real physical environment has become an essential problem. 
In order to more formally describe the problem, Markov Decision Processes (MDP) with tunable environment parameters are used in many works to model sim2real problems.
The objective of the sim2real is 
\begin{equation}
   \mathcal{J}(\theta) = \mathbb{E}_{\xi \sim p_{real}(\xi)}[ \mathbb{E}_{\tau \sim p_{real}(\tau)}[\sum_{t=0}^{T-1} \gamma^t r_{\xi}(s_t, a_t)| \pi^*_{sim}, \xi]]
   \label{eq_cmdp}
\end{equation}
where $\pi^*_{sim}$ is optimal policy trained in the simulator. We can obtain this policy by optimizing the following objective.
\begin{equation}
   \pi^*_{sim} = \arg \max_{\pi} \mathbb{E}_{\xi \sim p_{sim}(\xi)}[ \mathbb{E}_{\tau \sim p_{sim}(\tau)}[\sum_{t=0}^{T-1} \gamma^t r_{\xi}(s_t, a_t)| \pi, \xi]]
\end{equation}
Here the environment, also called domain, is characterized by the parameters $\xi$ which are assumed to be random variables distributed according to an unknown probability distribution $p_{real|sim}$. 
$s_t$ and $a_t$ are the state and action at time $t$, respectively. $r_{\xi}(\cdot)$ is the reward function. $\gamma$ is the discount factor.
The definitions of action space and state space of MDP are the same for environments with different parameters. 
The variability of the dynamics causes a distribution shift between the simulation data and the real data, thus degrading the performance of the trained policy during testing. 
% It is worth mentioning that this formula is very similar to contextual MDP(CMDP), which is usually used to solve the generalizability problem of reinforcement learning. 
% Further, MDP with tunable environment parameters can be seen as a subclass of the CMDP problem, 
% while the sim2real problem is a special case of the reinforcement learning generalizability problem.
For our proposed environment and tasks, in order to be able to study the sim2real problem more easily, 
we specify the possible parameters in the environment that cause the variability of the state transition function, 
and we describe each parameter in detail below.

\subsection{Robot Motion-related Parameters}
\subsubsection{Friction Coefficients}
Friction directly affects the acceleration and deceleration response processes of the robot, and friction coefficients are considered in many domain randomization works. 
With analysis of the mobile robot dynamics, the friction coefficients include two parameters: {the} sliding friction coefficient $f_{\bot}$ and {the} rolling friction coefficient $f_{\shortparallel}$. 
% Considering that the robot is in motion most of the time, here we have simplified the modeling by using the friction coefficient uniformly, 
% and the calculation process uses the dynamic friction calculation formula.

\subsubsection{Motor Character}
{The} motor character $C_T$ responds to the ability to generate torque, which directly affects the robot's acceleration and braking capabilities. 
Usually, the motor character is not constant and is related to the consumed time, the load, and the voltage.
In this paper, we only consider that the motor character can be approximated to a constant under normal working conditions.

\subsubsection{Controller Parameters}
We use the PID controller for the robot in the simulator and the physical robot, 
and the parameters $(k_p, k_i, k_d)$ are crucial to the controller. 
For a given robot, the PID control parameters are usually fixed. 
However, for different robots, due to the differences in actuators, the PID parameters are usually adjusted in order to achieve excellent performance for each robot. 
Therefore, the PID parameters may vary slightly from robot to robot.

\subsubsection{Robot Mass}
The mass $M$ of the robot affects the magnitude of the inertia and thus the control system. 
Therefore, many domain randomization methods in the literature take this parameter into account. 

\subsubsection{Rotational Inertia of the Wheel}
As can be seen from the dynamics model of the robot, the rotational inertia $\rho_w$ of the wheels affects the calculation of the velocity. 
Since the rotational inertia of the wheels on the physical robot is difficult to measure precisely, we also use it as an uncertain parameter.

\subsubsection{Control Response Latency}
The response latency $\zeta$ of the system can greatly affect the control effect of the robot, 
and there is also some work on reinforcement learning algorithms carried out specifically for system latency\cite{Ramstedt2019RealTimeRL}\cite{Ramstedt2020ReinforcementLW}. 
The system response latency usually includes several kinds, the first one is the node execution response latency due to limited computational resources 
during the simultaneous execution of multiple nodes or threads by the operating system, 
the second one is the response latency due to communication blockage between nodes and between the computer and the embedded system, 
and the other one is the execution response latency of the actuator itself. This delay parameter is usually not a constant value,  
but a random variable due to the chance of occurrence of the above-mentioned cases.
In this paper, we no longer distinguish the causes of delay, but divide the control delay into four parts from the results of the delay: 
longitudinal velocity delay $\zeta_{v_x}$, lateral velocity delay $\zeta_{v_y}$, angular velocity delay $\zeta_{v_w}$ and shooting control delay $\zeta_{s}$. In the simulation environment, we simulate delays in real physical systems by delaying the deployment time of the control.

\subsection{Sensor-related Parameters}
\subsubsection{LiDAR Noise Parameters}
In the previous sections, we analyze the sources of LiDAR noise and build the LiDAR noise model using a Gaussian distribution. 
Where we assume that the mean $\mu_l$ of the Gaussian distribution is 0 and the standard deviation $\sigma_l$ is related to the LiDAR measurement performance.

\subsubsection{LiDAR Anomaly Parameters}
LiDAR data anomalies {may be} caused by sensor anomalies, but a more likely cause is {that} the measurement of {the} laser reflectivity of special object materials is too low or {that}
some special laser incidence angles, resulting in LiDAR not receiving reflected data. 
This latter phenomenon is related to the layout of the arena and the {materials} of the environment. 
In the previous sections, we employ the Poisson distribution to build the LiDAR anomaly model, 
so the parameter to be determined is $\lambda$ in the Poisson distribution.
\subsection{Communication Protocol}
{
We have defined communication protocols between the 4 modules of parameter manager, simulation system, physical system, and policy to achieve the cooperation between the modules. There are two types of communication protocols here: one involves adjustable parameters; Another type involves policy interfaces.
}
\subsubsection{Adjustable Parameters}
{
In the previous two subsections, we describe in detail the adjustable parameters of robot motion and sensors. The parameter manager controls the properties of the simulation environment during policy training by setting these parameters and passing them to the simulation system, enabling the policy to achieve different generalization performance.
}

\subsubsection{Policy Interfaces}
{
This interface involves the states of policy receiving and the actions of output. The purpose of this interface is to separate policy algorithm development from task deployment, in order to reduce the additional difficulties caused by considering task deployment during the algorithm development. On the other hand, this part of the communication protocol includes preprocessing of states in simulation and physical systems, as well as post-processing of policy output actions, thereby reducing the difficulty of policy deployment.
}
\subsection{Baselines}
At present, there are many sim2real works aiming at the dynamics difference between simulation and real environment such as system identification\cite{Punjani2015DeepLH}\cite{Zhu2017FastMI}, domain randomization\cite{Peng2017SimtoRealTO}, 
adaptive domain randomization\cite{Chebotar2018ClosingTS}\cite{Mehta2019ActiveDR}, etc. In this paper, we choose 5 methods to build the baselines of our proposed tasks. 

\emph{State Space and Action Space: }
{
In this platform, we provide a variety of sensor data that can be used to construct the state space for reinforcement learning. Images, LiDAR point clouds, and robot pose and velocity related information can be used. In the baseline algorithm of this article, the state space consists of 69 dimensional data including LiDAR data, robot pose, robot velocity, and goal pose. The action space is composed of the speed control of the robot, including lateral velocity, longitudinal velocity, and rotational angular velocity in the robot coordinate system.
}

Its reward function is set as follows
\begin{equation}
   r_n = \begin{cases}
      40, & \text{ IF goal activated} \\
      d_t - d_{t-1} + \frac{|\Delta \theta_t| - |\Delta \theta_{t-1}|}{2} \\
      - {\beta} \min(\exp(\frac{n_s}{4000}-5), 1) - 0.1. & \text{ELSE} \\
   \end{cases}
\end{equation}
Here $d_t$ is the distance from the robot to the goal at the timestep $t$. 
$\Delta \theta_t$ is the angle difference between the robot orientation and the line to the goal. 
$n_s$ and {$\beta$} are the training iteration and the weight of the collision, respectively.
{
$\beta$ will affect the sensitivity of the learned policy to collisions during the training. When the value is relatively large, the policy will be very conservative and affect the exploratory behavior during the policy learning.}
We use Soft-Actor-Critic (SAC) \cite{Haarnoja2018SoftAO} algorithm to train the navigation policy in the simulation, and evaluate the sim2real performance by transferring the trained policy to physical robots. 
In order to reduce performance degradation, we have adopted the following 5 sim2real methods in the training process.
\subsubsection{None}
% Compared with the sim2real method, we use the default parameters of the simulator to train, and directly transfer the trained policy to the physical robot.
We implement a {naive} transfer method that does not rely on any real robot a priori or data, train the policy using the default parameters of the simulator, 
and then transfer it to the physical robot and use it as a comparison to other sim2real methods. The method is shown in $(a)$ of Fig. \ref{fig_baselines}.

\subsubsection{Uniform Domain Randomization(UDR)\cite{Peng2017SimtoRealTO}}
% Because of its simplicity and effectiveness, this method has been widely used in many current works. However, how to choose a reasonable random interval is the key to this method.
According to the literature\cite{Peng2017SimtoRealTO}, we reproduce the uniform domain randomization method. In contrast to using the default parameters, 
we set a reasonable random sampling interval for each parameter based on the engineering experience. 
During the training process, the simulator acquires a set of parameters through the parameter generator to collect robot trajectory data. 
The method is shown in $(b)$ of Fig. \ref{fig_baselines}.

\subsubsection{DROID\cite{Tsai2021DROIDMT}}
% This method is an offline adaptive domain randomization method. The difference of SimOpt is that it does not interact with the physical system, only needs offline dataset of physical system, 
% and uses CMA-ES\cite{Hansen2001CompletelyDS} to optimize the dynamic parameter distribution.
In contrast to UDR, this method draws on offline expert data to optimize the simulator parameters by reducing the error between the simulated and real trajectories. 
In the implementation of this paper, we minimize the $l2$-norm loss of the state, including the pose and velocity, by means of the CMA-ES\cite{Hansen2001CompletelyDS}. The method is shown in $(c)$ of Fig. \ref{fig_baselines}.

\subsubsection{SimOpt\cite{Chebotar2018ClosingTS}}
% This method is an online adaptive domain randomization method. In the training process, it is necessary to use the trained policy to collect data in the real environment, 
% and then play back in the simulator, and use the relative entropy policy search\cite{Peters2010RelativeEP} to optimize the dynamic parameters to reduce the difference between the real data and the virtual data.
SimOpt defines a discrepancy metric between real and simulated trajectories and uses relative entropy policy search\cite{Peters2010RelativeEP} to optimize the simulator parameter distribution. 
Unlike DROID, this method requires interaction with the real robot to collect real data and continuously update the simulator parameters. We still consider only robot poses and velocities when reproducing the trajectory discrepancy metrics. 
The method is shown in $(d)$ of Fig. \ref{fig_baselines}.

\subsubsection{Ground Action Transformation(GAT)\cite{Hanna2021GroundedAT}}
% This method does not need to tune the parameters of the simulator, but constructs an action transformer outside the simulator, 
% so that the next state in the simulation environment is the same as in the real data under the same current state and action.
Unlike the previous domain randomization methods, GAT does not require modifications to the simulator parameters, 
but instead constructs an action transformer outside the simulator to make the next state in the simulation environment the same as in the real environment with the same current state and action. 
Again, the method requires interaction with the real robot to collect data to build the action transformer. The method is shown in $(e)$ of Fig. \ref{fig_baselines}.

\begin{figure}
   \hspace{0.5cm}
	\centering  
   \includegraphics[width=0.48\textwidth]{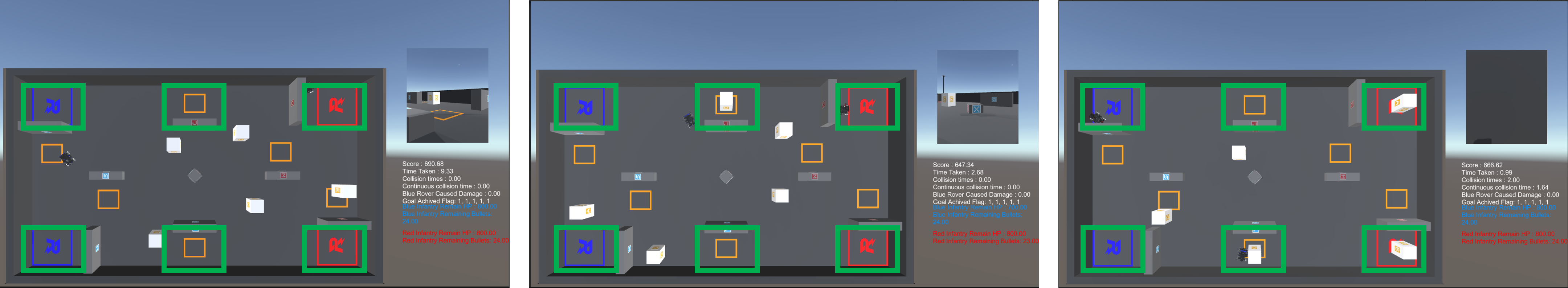} 
   
	\caption{Blocking zones and evaluation scenarios. The green squares are the blocking zones. From left to right are the samples of the evaluation scenarios from Level 1, Level 2, and Level 3.}
   \label{fig_scenarios}
   \vspace{-0.5cm}
\end{figure}

\section{Evaluation Scenarios and metrics}
% For the competition, we have defined a composite evaluation metric to assess the performance of the intelligences by calculating the score of the algorithm in each game. 
% However, it is difficult for us to analyze the more specific performance and characteristics of the intelligences by this one comprehensive evaluation score alone. 
% Therefore, we have set different evaluation metrics for the two tasks of navigation and confrontation
In order to evaluate the performance of different methods, evaluation scenarios with different levels are defined. 
Moreover, we set corresponding evaluation metrics for different tasks.
\subsection{Evaluation Scenarios}
In order to evaluate the navigation ability of the agent comprehensively, we design 3 levels of evaluation scenarios based on the location distribution of the goals. 
Specifically, we designate the blocking zones on the arena, as shown in Fig. \ref{fig_scenarios}. 
{
The blocking zone is a fixed square area with a side length of 1 $m$ on the map. At least two edges of the zone are occupied by obstacles.
These zones are usually blocked by surrounding obstacles, 
and the agent in these zones needs to bypass the obstacles to activate the goals. 
}
We set up 3 different levels according to the number of goals within the blocking zones.
\begin{itemize}
   \item \textbf{Level 1}: All goals are outside the blocking zones;
   \item \textbf{Level 2}: There is 1 goal randomly generated within the blocking zones;  
   \item \textbf{Level 3}: There are 3 goals randomly generated within the blocking zones.
\end{itemize}
\subsection{Navigation Task}
% Currently, many works studying navigation tasks follows the evaluation metrics of the article. One of the metrics is the SR, 
SR is one of the most commonly used evaluation metrics in navigation tasks. 
It is the percentage of successfully reaching the goals over multiple trials, and for the task in this paper, it is the percentage of successfully activating the goals. 
\begin{equation}
   \text{SR} = \frac{1}{N} \sum^N_{i=1} \frac{\mathcal{N}^a_i}{\mathcal{N}^g_i}
\end{equation}
Here $\mathcal{N}^a_i$ is the number of activated goals in the trial $i$. $\mathcal{N}^g_i = 5$ is the total number of goals in each trial.
$N$ is the number of trials.
Another one is the SPL.
\begin{equation}
   \text{SPL} = \frac{1}{N} \sum^N_{i=1} \mathcal{S}_i \frac{l_i}{\max(p_i, l_i)}
\end{equation}
% $SPL = S \frac{l}{\max (l, p)} $
where $l_i$ is the shortest path from the starting point to the goal. $p_i$ is the length of the path traveled by the agent. 
$\mathcal{S}$ is an indication of successful activation, and $\mathcal{S}_i=1$ means the goal is successfully activated in the trial $i$.
However, for physical robots, the two aforementioned metrics do not assess the safety of the agent during navigation. 
Therefore, we define the SaFety weighted by Path Length (SFPL), which is calculated by following 
\begin{equation}
   \text{SFPL} = \frac{1}{N} \sum^N_{i=1} \frac{\sum^T_{t=1} (1 - \alpha_{i,t})\delta_{i, t}}{p_i}
\end{equation}
where $\alpha_{i,t}$ is the collision identifier. If the robot collides in time $t$ of the trial $i$, $\alpha_{i,t}$ is 1, otherwise is 0. $\delta_{i,t}$ indicates the distance the robot moves in time $t$.
Compared with the number of collisions, this metric can more reasonably evaluate the safety of a robot sliding against the wall.
In addition, we calculate the average velocity of each trajectory during the episode, 
which is used to measure the efficiency of the agent during navigation.
\begin{equation}
   \text{Velocity} = \frac{1}{N}\sum_{i=1}^N \frac{1}{T}\sum_{t=1}^T v_{i,t}
\end{equation}
where $v_{i, t}$ is the velocity of the robot in the time $t$ of the trial $i$. 
Besides the metrics about the navigation performance, in order to compare the difference between the simulated trajectory and the real trajectory, 
we employ the Wasserstein distance and design a metric named TrajGap for evaluating the trajectory distribution gap of the policy in the simulator and the real environments.
\begin{equation}
   \text{TrajGap} = \frac{1}{N}\sum_{i=1}^{N} \inf_{\kappa \sim \Pi} \int_{x \in S_{sim}} \int_{y \in S_{real}} \kappa(x, y)||x-y|| dx dy
\end{equation}
Here $\Pi$ is the joint distribution of simulated and real data. $S$ is the state of the trajectory which includes the robot's velocity and pose.

\section{Experiments}
In the previous sections, we construct the robot dynamics model and the sensor simulation model. 
In the experimental section, we first validate the robot model and the sensor models. Then we analyze the safety metrics to evaluate the navigation trajectories, 
and the baselines are evaluated in evaluation scenarios for navigation tasks.

\subsection{Robot Dynamic Model and Sensor Simulation Model}

% In the previous session, we constructed the dynamic model of the robot, especially for the mecnum wheel, and established the friction simulation model. 
% In addition, we also build a noise model for LiDAR. Therefore, here we will mainly validate the dynamic model of the robot and the LiDAR noise model.
In the previous sections, we construct the dynamics model of the robot and build a complex friction model of the Mecanum wheel structure. 
In addition, we also build the noise model and the anomaly model for the LiDAR data. Therefore, in this section, we focus on answering two questions.
\begin{itemize}
   \item \textbf{Q1}: Will it be better to use the friction model to simulate the dynamic response of the real robot?
   \item \textbf{Q2}: Can the noise model and the anomaly model simulate the real LiDAR data?
\end{itemize}

\begin{figure}
   \hspace{0.5cm}
	\centering  
   \includegraphics[width=0.4\textwidth]{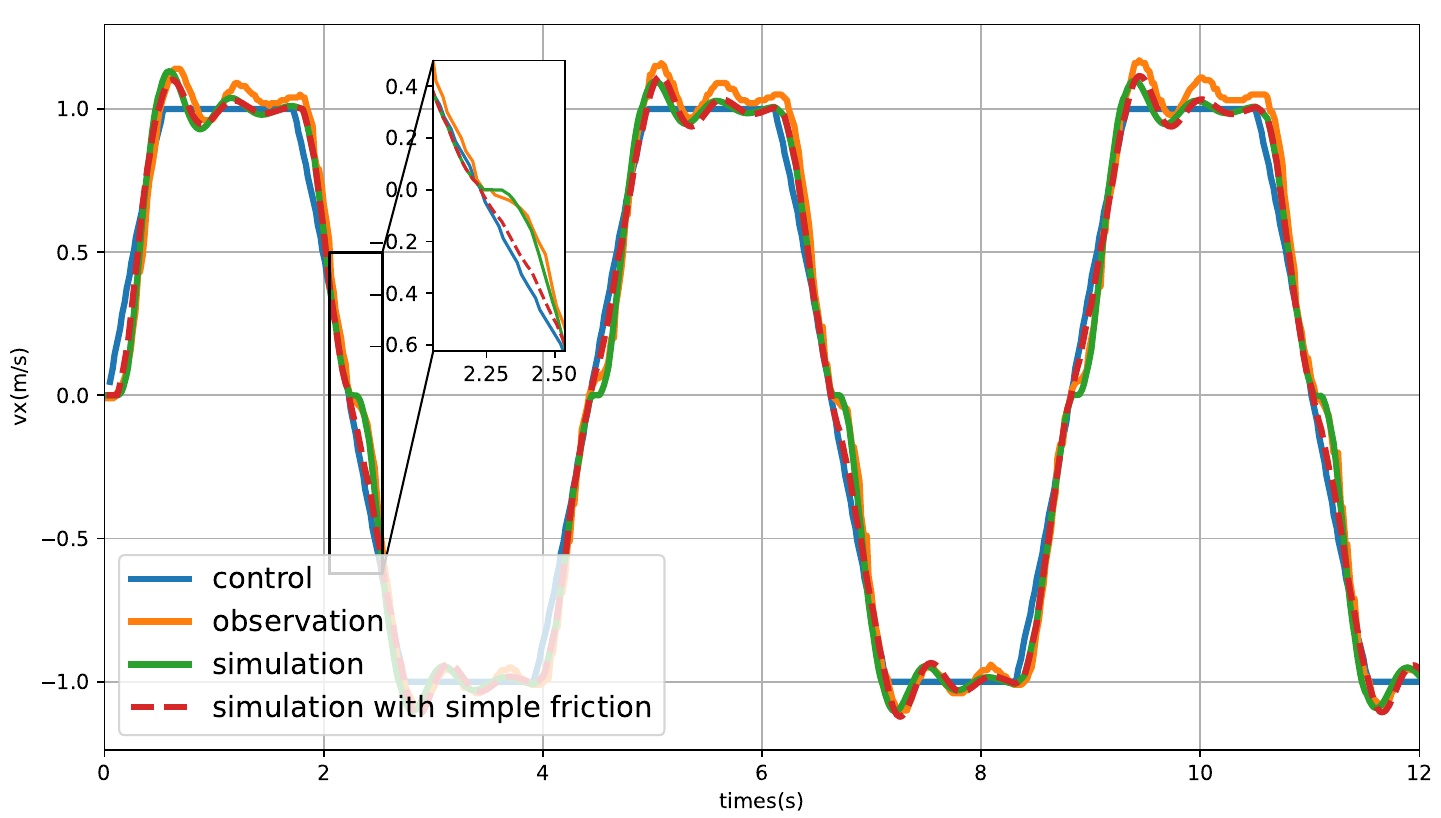} 
   
	\caption{Input control and longitudinal speed response of the robot.}
   \label{fig_dynamic}
   \vspace{-0.5cm}
\end{figure}

% \subsubsection{Will it be better to use the friction model to simulate the dynamic response of the real robot?}
\emph{1) Will it be better to use the friction model to simulate the dynamic response of the real robot?}
We set up a set of control sequences on the real robot and collect the velocity of the robot at each moment. 
We also apply the sequence to the simulation environment and collect the simulated velocity. The parameters of the simulation are shown in Table \ref{table_sim_param}
For comparison, we also construct a common simplified friction model. The experimental results are shown in Fig. \ref{fig_dynamic}.

From the results in the figure, we can see that our friction model can better simulate the speed response of the robot than the simplified friction model. 
Especially when the speed direction changes due to the combined effect of rolling and sliding friction, 
the simplified friction model makes it difficult to simulate the non-linear change of speed. 
\begin{table}
   % \vspace{-0.5cm}
   \hspace{0cm}
   \center
   \caption{Parameters of the simulation environment.}
   \begin{tabular}{c c c c c}
      \hlinew{1pt}
      $f_{\bot}$, $f_{\shortparallel}$& $C_T$& $(k_p, k_i, k_d)$ & $M$ & $\rho_w$ \\
      \hline
      (0.069, 0.082)  & 0.375 & (12.5, 0.0, 2.0) & 0.45 & 0.096   \\
      \hline
      &\multicolumn{3}{c}{($\zeta_{v_x}$, $\zeta_{v_y}$, $\zeta_{v_w}$, $\zeta_{s}$)}&  \\
      \hline
      & \multicolumn{3}{c}{(0.02, 0.3, 0.04, 0.72)} & \\
      \hlinew{1pt}
   \end{tabular}
   \label{table_sim_param}
   \vspace{-0.5cm}
\end{table}

\begin{figure}
   \hspace{0.5cm}
	\centering  
   \includegraphics[width=0.49\textwidth]{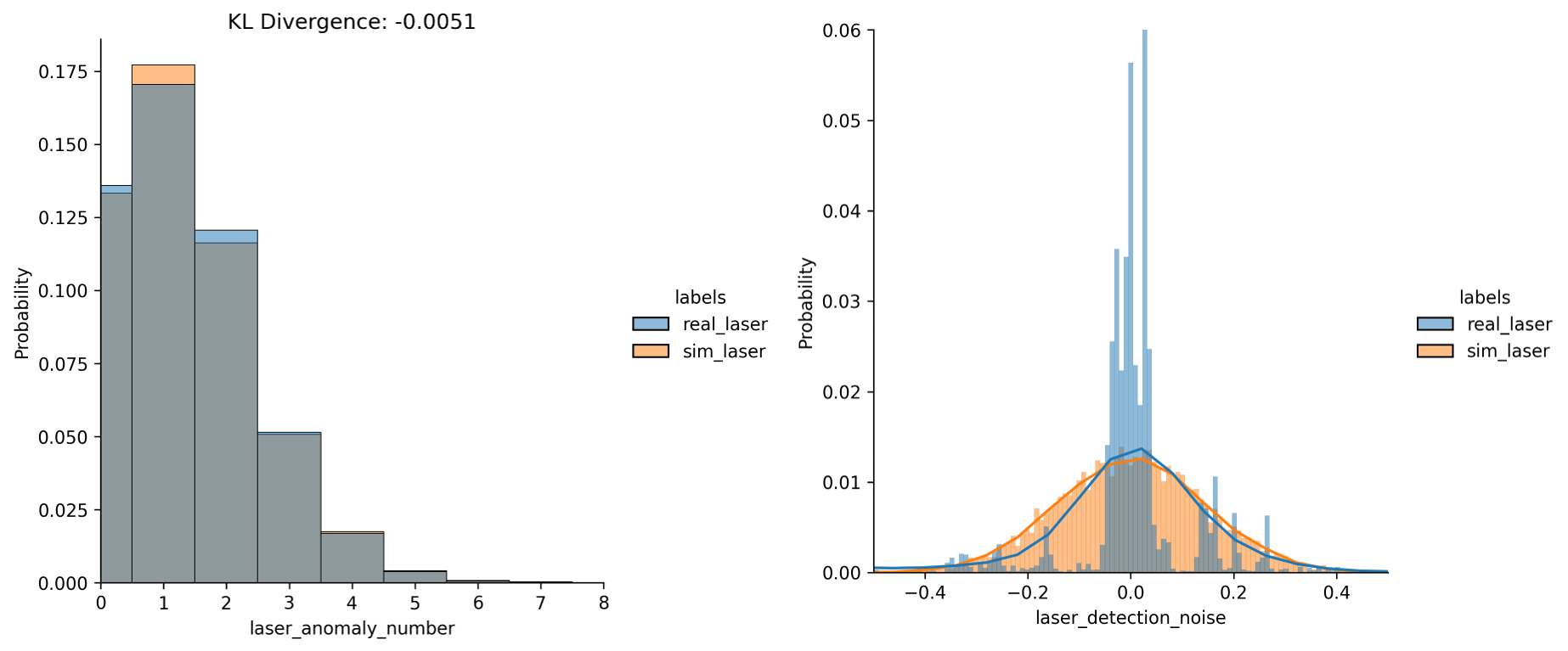} 
   
	\caption{Data distributions of simulation LiDAR and real LiDAR. 
   The left is the distribution of the abnormal points per frame, and the right is the distribution of LiDAR ranging noise data.}
   \label{fig_laser_anomaly}
   \vspace{-0.5cm}
\end{figure}

% \subsubsection{Can the noise model and the anomaly model simulate the real LiDAR data?}
\emph{2) Can the noise model and the anomaly model simulate the real LiDAR data?}
To verify the LiDAR anomaly data model, we collect 1000 frames of LiDAR data {at} different locations in the real environment. 
The left part of Fig. \ref{fig_laser_anomaly} shows the real LiDAR anomaly data distribution and the distribution of data sampled from the anomaly model. 
It can be seen that the real data distribution is very consistent with the simulated data distribution, 
and the KL divergence between the two data distributions is very small $(-0.0051)$.

To verify the LiDAR noise model, we randomly select five locations in the real environment and collected 50 frames (5 seconds) of data at each location. 
We calculate the measurement noise of LiDAR by using the average of 50 frames as the true measurement value. 
The right part of Fig. \ref{fig_laser_anomaly} shows the real LiDAR noise data distribution and the data distribution of the noise model. 
From the results in the figure, it can be seen that the real LiDAR noise dataset is within the range of $[-0.05, 0.05]$. 
Within this range, our model can better simulate the data distribution of noise. 
{From the right part of Fig. \ref{fig_laser_anomaly}, it seems that using a Gaussian mixture model can better fit the data distribution of LiDAR measurement errors. However, if more data is collected, the center position of the dwarf peak may change, making it difficult to accurately estimate the mean of Gaussian mixture distribution. In addition, the purpose of modeling the LiDAR error is to adapt the policy to sensor measurement errors during the training. On the one hand, we need to determine a reasonable range of errors to consider various error situations during training. On the other hand, we also need to avoid overfitting and conservatism of the policy caused by selecting too narrow or too wide errors. Therefore, we use Gaussian models instead of Gaussian mixture models. Compared to Gaussian mixture models, the mean and variance of Gaussian are easy to determine and have good generalization, and they can cover the error range well. In addition, the sampling is relatively simple and computationally efficient, which is beneficial for the training.}

\begin{table}
   % \vspace{-0.5cm}
   \hspace{0cm}
   \center
   \caption{Comparison of the different safety metrics.}
   \begin{tabular}{c c c}
      \hlinew{1pt}
      Trajectories & SFPL$\uparrow$ & Collision Number$\downarrow$ \\
      \hline
      Trajectory1  & 0.854 & 20 \\
      Trajectory2  & 0.842 & 13 \\
      Trajectory3  & 0.770 & 14 \\
      Trajectory4  & 0.726 & 12 \\
      \hlinew{1pt}
   \end{tabular}
   \label{table_collision}
   \vspace{-0.5cm}
\end{table}

\begin{figure}
   \hspace{0.5cm}
	\centering  
   \includegraphics[width=0.49\textwidth]{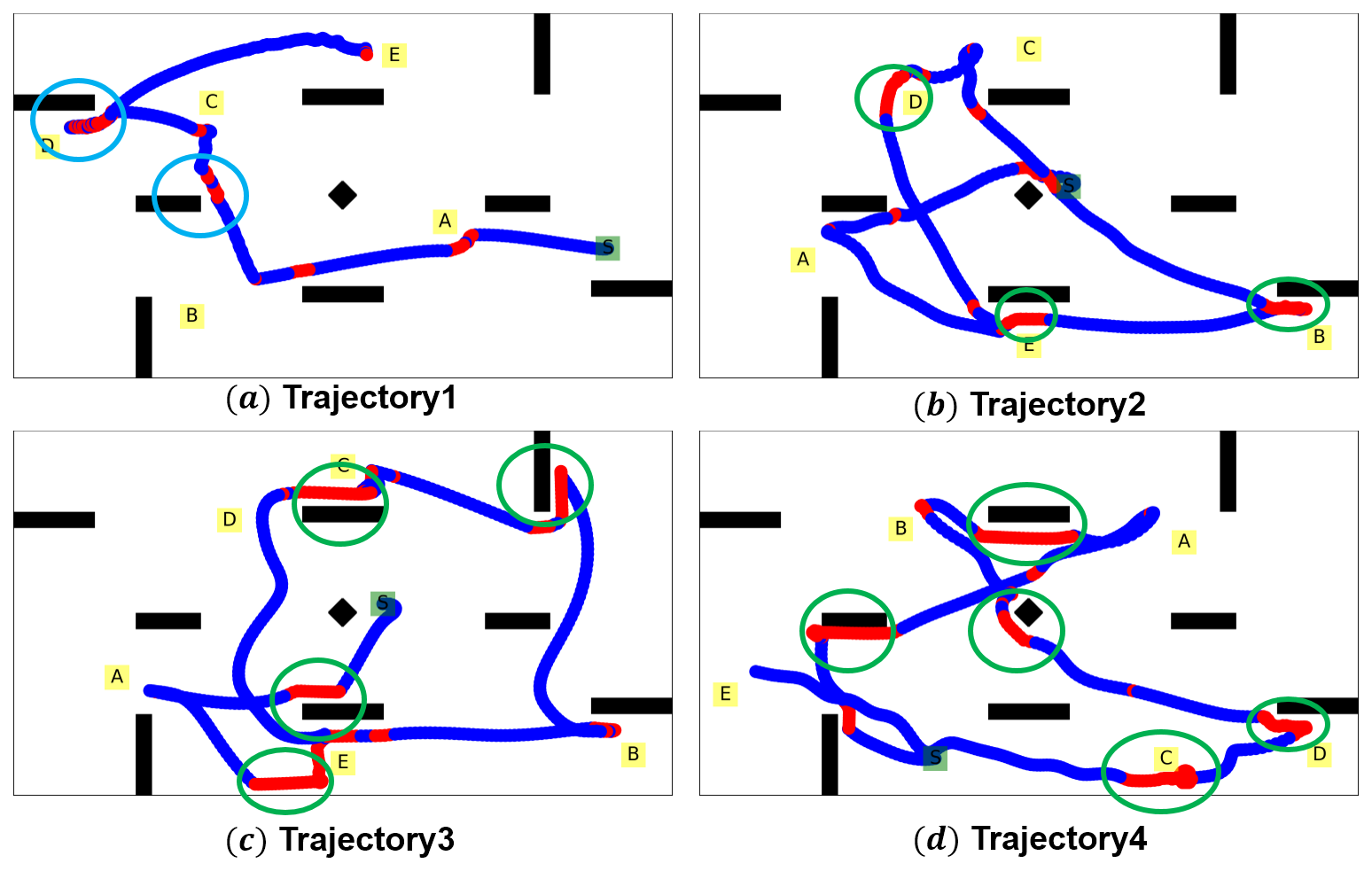} 
   
	\caption{The collision visualization of the different trajectories. 
   Blue indicates a safe trajectory and red indicates the collision between the agent and the obstacle.
   The cyan circle indicates that the robot continuously collides with the obstacle, which happens in Trajectory1$(a)$. 
   The green circle indicates that the robot slides against the obstacles, which happens in Trajectory2$(b)$, Trajectory3$(c)$ and Trajectory4$(d)$.}
   \label{fig_collision}
   \vspace{-0.5cm}
\end{figure}

\begin{table}
   \hspace{0cm}
   \center
   \caption{Comparison of the different data augmentation methods on the navigation scenarios.}
     \begin{tabular}{c c c c c}
      \hlinew{1pt}
      % & \multicolumn{4}{c}{RMMS}\\
   %   \cline{2-5}
      Scenarios & Methods & SR$\uparrow$ & SPL$\uparrow$ & SFPL$\uparrow$ \\
      \hline
      \multirow{4}{*}{Level 1} & None & 0.91 & 0.83 & 0.87  \\
                           %  \cline{2-5}
                            & Noise & 0.95 & 0.87 & 0.88 \\
                           %  \cline{2-5}
                            & Shift & 0.95 & 0.87 & 0.85 \\
                           %  \cline{2-5}
                            & Noise+Shift & 0.93 & 0.84 & 0.84 \\
      \hline
      \multirow{4}{*}{Level 2} & None & 0.83 & 0.79 & 0.85 \\
                           %  \cline{2-5}
                            & Noise & 0.91 & 0.84 & 0.83 \\
                           %  \cline{2-5}
                            & Shift & 0.74 & 0.77 & 0.82 \\
                           %  \cline{2-5}
                            & Noise+Shift & 0.77 & 0.77 & 0.83 \\
      \hline
      \multirow{4}{*}{Level 3} & None & 0.68 & 0.71 & 0.81 \\
                           %  \cline{2-5}
                            & Noise & 0.87 & 0.80 & 0.83 \\
                           %  \cline{2-5}
                            & Shift & 0.65 & 0.69 & 0.81 \\
                           %  \cline{2-5}
                            & Noise+Shift & 0.63 & 0.72 & 0.79 \\
     \hlinew{1pt}
     \end{tabular}%
   \label{table_data_aug}
   \vspace{-0.5cm}
\end{table}%

\begin{table*}
   \hspace{0cm}
   \center
   \caption{The ablation study on the robot dynamics model and LiDAR model.}
   \begin{tabular}{c c c c c c c c c c c c c}
      \hlinew{1pt}
                & \multicolumn{3}{c}{{SR$\uparrow$}} & \multicolumn{3}{c}{{SPL$\uparrow$}}  & \multicolumn{3}{c}{{SFPL$\uparrow$}} & \multicolumn{3}{c}{{Velocity$\uparrow$}} \\
      \hline
      {Methods} & {sim}  &{real} & {$|$gap$|\downarrow$}               & {sim}  &{real} & {$|$gap$|\downarrow$}        & {sim} & {real} & {$|$gap$|\downarrow$ }            & {sim} & {real} & {$|$gap$|\downarrow$}    \\
      \hline
      {None} & {1.0} & {0.90} & {0.10} & {0.96} & {0.70} & {0.26} & {0.85} & {0.38} & {0.47} & {1.0} & {0.81} & {0.19}  \\
      {w. RD}  & {1.0} &{0.85} & {0.15} & {0.94} & {0.65} & {0.29} & {0.86} & {0.40} & {0.46} & {1.08} & {0.85} & {0.23}  \\
      {w. LiDAR} & {1.0} & {1.0} & {0.0} & {0.92} & {0.80} & {0.12} & {0.81} & {0.49} & {0.32} & {1.0} & {0.91} & {0.09}  \\
      {w. ALL} & {1.0} & {1.0} & {0.0} & {0.94} & {0.83} & {0.11} & {0.80} & {0.51} & {0.29} & {1.10} & {0.93} & {0.17}  \\ 
      \hlinew{1pt}
   \end{tabular}
   \label{table_abulation}
   % \vspace{-0.5cm}
\end{table*}

\subsection{Comparison of Safety Metrics}
Reasonable safety metrics are crucial to evaluating the policy on the physical robot. The number of collisions is a common safety evaluation metric in the literature. 
In this section, we focus on the following question:
\begin{itemize}
   \item \textbf{Q}: Is SFPL a more reasonable metric to evaluate trajectory safety than the number of collisions?
\end{itemize}

{We evaluated different evaluation metrics on 4 typical collision scenarios, as shown in Fig. 8. The scenario in (a) of Fig. \ref{fig_collision} shows the discrete collisions. The other parts include both discrete collision and continuous collision scenarios.}
We analyze the collected navigation trajectories to compare the rationality of SFPL and collision {numbers} for evaluating safety. 
The experimental results are shown in Table \ref{table_collision}.
It can be seen that the assessment results of the trajectory by SFPL and the number of collisions are not consistent. 
While the SFPL of Trajectory1 and Trajectory2 are relatively close, the number of collisions differs greatly. The number of collisions between Trajectory2 and Trajectory3 is close, 
but SFPL is quite different. The evaluation of the two metrics for Trajectory3 and Trajectory4 is consistent. We visualize the trajectories, as shown in Fig. \ref{fig_collision}. 
It can be seen from the figure that, compared with Trajectory2, Trajectory1 has continuous collisions, which will result in a large number of collisions. 
But on the whole, the safety of the two trajectories is similar, and Trajectory1 is safer. In Trajectory3 and Trajectory4, the agent slides against the obstacle. 
Since the length of the sliding path does not affect the number of collisions, the number of collisions between Trajectory3 and Trajectory4 is similar to Trajectory2. 
However, from the perspective of trajectory distribution, Trajectory2 is safer than Trajectory3 and Trajectory4. Therefore, we can conclude that, compared with the collision number, 
SFPL can more reasonably evaluate the safety of the trajectory.

\begin{table*}
   \hspace{0cm}
   \center
   \caption{Comparison of the baselines on the different levels scenarios.}
   \begin{tabular}{c c c c c c c c c c c c c c c}
      \hlinew{1pt}
                &          & \multicolumn{3}{c}{SR$\uparrow$} & \multicolumn{3}{c}{SPL$\uparrow$}  & \multicolumn{3}{c}{SFPL$\uparrow$} & \multicolumn{3}{c}{Velocity$\uparrow$} & TrajGap$\downarrow$ \\
      \hline
      Scenarios & Methods & sim  & real & $|$gap$|\downarrow$               & sim  & real & $|$gap$|\downarrow$        & sim & real & $|$gap$|\downarrow$             & sim & real & $|$gap$|\downarrow$            &   \\
      \hline
      \multirow{5}{*}{Level 1}& None & 1.0 & 1.0 & 0.0 & 0.94 & 0.83 & 0.11 & 0.80 & 0.51 & 0.29 & 1.10 & 0.93 & 0.17 & 1.77 \\
                             & UDR  & 1.0 & 1.0 & 0.0 & 0.86 & 0.79 & 0.07 & 0.75 & 0.75 & 0.0 & 1.16 & 0.99 & 0.17 & 1.90 \\
                             & DROID & 1.0 & 1.0 & 0.0 & 0.83 & 0.79 & 0.04 & 0.77 & 0.75 & 0.02 & 1.0 & 0.81 & 0.19 & 1.89 \\
                             & SimOpt & 1.0 & 1.0 & 0.0 & 0.95 & 0.92 & 0.03 & 0.88 & 0.76 & 0.12 & 1.05 & 1.00 & 0.05 & 0.59 \\ 
                             & GAT  & 1.0 & 1.0 & 0.0 & 0.91 & 0.83 & 0.08 & 0.85 & 0.62 & 0.23 & 0.96 & 0.94 & 0.02 & 1.76 \\
      \hline
      \multirow{5}{*}{Level 2} & None & 1.0 & 1.0 & 0.0 & 0.83 & 0.65 & 0.17 & 0.69 & 0.49 & 0.20 & 0.92 & 0.84 & 0.08 & 1.97 \\
                              & UDR  & 1.0 & 1.0 & 0.0 & 0.81 & 0.84 & 0.03 & 0.68 & 0.58 & 0.10 & 0.97 & 0.88 & 0.10 & 1.56 \\
                              & DROID & 1.0 & 1.0 & 0.0 & 0.83 & 0.78 & 0.05 & 0.67 & 0.68 & 0.01 & 0.91 & 0.99 & 0.08 & 1.93 \\
                              & SimOpt & 1.0 & 1.0 & 0.0 & 0.94 & 0.81 & 0.13 & 0.88 & 0.67 & 0.21 & 1.03 & 0.96 & 0.07 & 1.76 \\
                              & GAT  & 1.0 & 1.0 & 0.0 & 0.85 & 0.74 & 0.11 & 0.78 & 0.58 & 0.20 & 0.94 & 0.90 & 0.04 & 1.06 \\
      \hline
      \multirow{5}{*}{Level 3} & None & 1.0 & 1.0 & 0.0 & 0.83 & 0.64 & 0.19 & 0.74 &  0.59 & 0.15 & 1.04 & 0.86 & 0.18 & 3.07 \\
                            & UDR  & 1.0 & 1.0 & 0.0 & 0.75 & 0.85 & 0.10 & 0.73 & 0.63 & 0.10 & 1.13 & 0.97 & 0.16 & 2.97 \\ 
                            & DROID & 1.0 & 1.0 & 0.0 & 0.87 & 0.88 & 0.01 & 0.79 & 0.70 & 0.09 & 1.18 & 1.01 & 0.17 & 1.21 \\
                            & SimOpt & 1.0 & 1.0 & 0.0 & 0.89 & 0.79 & 0.10 & 0.86 & 0.84 & 0.02 & 1.00 & 0.93 & 0.07 & 1.71 \\
                            & GAT   & 1.0 & 1.0 & 0.0 & 0.79 & 0.61 & 0.18 & 0.71 & 0.59 & 0.12 & 0.80 & 0.80 & 0.0 & 1.27 \\
      \hlinew{1pt}
   \end{tabular}
   \label{table_scenario_all}
   % \vspace{-0.5cm}
\end{table*}
\begin{figure*}
   \hspace{0.5cm}
	\centering  %图片全局居中
	% \subfigbottomskip=1pt %两行子图之间的行间距
	% \subfigcapskip=-3pt %设置子图与子标题之间的距离
   \includegraphics[width=1.0\textwidth]{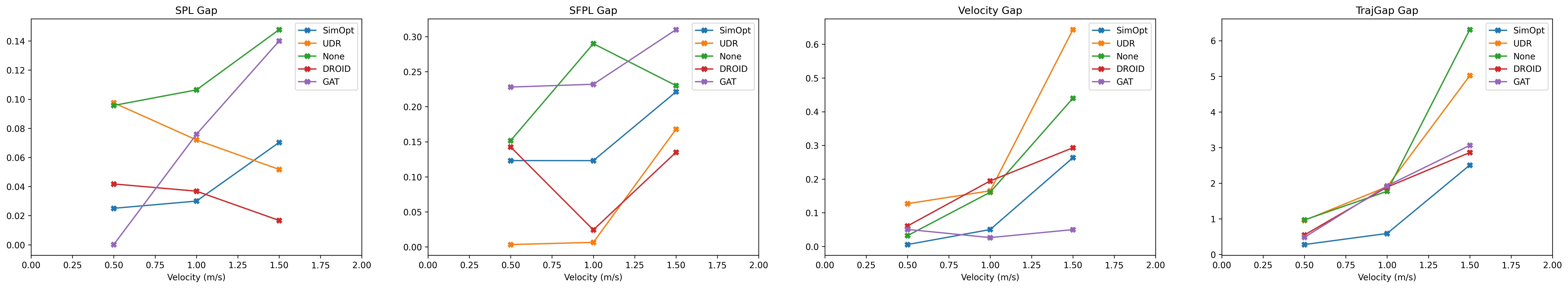} 
   
	\caption{The sim2real gaps of different metrics with various maximum velocities.}
   \label{fig_speed_results}
   \vspace{-0.5cm}
\end{figure*}

\subsection{Comparison of Baselines on the Navigation Task}
% In this session, we first discuss the effects of different data augmentation methods on the performance of the algorithm during the algorithm training process. 
% Then we focus on the two aspects of the navigation task that affect the algorithm sim2real: 1) the difficulty of the navigation task and 2) the maximum navigation speed.
In this section, we focus on evaluating the performance changes of the baselines transferring from the simulation to the real in different navigation scenarios. 
Specifically, we will answer 3 questions.
\begin{itemize}
   \item \textbf{Q1}: Does the data augmentation commonly used during navigation training still work to improve the generalization of the method?
   \item {\textbf{Q2}: What impact do robot dynamics model and the LiDAR model have on sim2real?}
   \item \textbf{Q3}: How are the performance of the  sim2real baselines in different evaluation scenarios?
   \item \textbf{Q4}: What will happen on the performance of sim2real baselines if the robot moving speed changes greatly?
\end{itemize}

\subsubsection{Effect of different data augmentation methods during training (\textbf{Q1})}
Data augmentation is an important way to improve the generalization of deep reinforcement learning algorithms. 
We consider 3 different data augmentation methods for the robot pose: pure noise, pure bias, and noise bias combination. 
% For the position and posture of the robot, three different data augmentation methods are considered: pure noise, pure bias and noise bias combination. 
Among them, the noise is to simulate the localization error of the real robot owing to the sensor noise, 
and the bias is to simulate the localization offset of the real robot when the wheels slip or collide with the obstacles.

Table \ref{table_data_aug} shows the results of the policy trained by SAC\cite{Haarnoja2018SoftAO} after 1000 iterations. 
These results are the averages {of} 20 trials in each scenario. 
It can be seen from the experimental results that the data augmentation methods improve the test performance of the policy in Level 1 scenarios. 
For Level 2 and Level 3 scenarios, {the combination of pure bias and noise bias degrades the algorithm's performance}
since the disturbance of noise and bias to the localization makes the task more difficult. 
Navigation with localization offset is a challenging task compared to localization noise. 
Overall, the results in the table show that appropriately increasing the difficulty of navigation tasks can improve the generalization of the policy and that {overly difficult} tasks may result in degraded performance.

\subsubsection{The ablation study on the robot dynamics model and the LiDAR model (\textbf{Q2})}
{
In this section, we analyze the impact of the proposed robot dynamics model and LiDAR model on sim2real. We  use the learned policy with removing the robot dynamics model and LiDAR model as the baseline. We separately add the dynamic model and the sensor model to analyze the impact of different modules on the learned policies. In the simulation without the dynamics model, the speed of the robot is the same as the control input. In the simulation without the LiDAR model, the LiDAR measurement data is a true value without any noise or disturbance. We conduct 20 trails in Level 1 scenario with a maximum speed of 1 $m/s$ for the robot. The results are shown in Table \ref{table_abulation}. Among them, ``None" indicates that neither of the above two models is used. ``w. RD", ``w. LiDAR", and ``w. ALL" represents the use of the dynamic model, the LiDAR model, and both models, respectively. It can be seen that without using the dynamic model and the LiDAR model, the robot policy is very prone to overfitting, resulting in a significant decrease in performance in the real environment, although it may improve the performance of the policy in the simulation environment. Moreover, we can see that compared to the dynamics model, the LiDAR model plays a more important role in the sim2real  of robot navigation.
}

\subsubsection{The effect of evaluation scenarios on sim2real transfer performance (\textbf{Q3})}
% In the previous section, we set up three levels of navigation tasks based on the location of the targets. 
In this section, we compare the performance of the baselines and discuss the impact of different levels on sim2real in the 3 test scenarios.

The performance of the baselines at different levels of navigation tasks is detailed in Table \ref{table_scenario_all}. 
The policies used for evaluation are trained with 2000 iterations. The maximum speed of the policies is $1.0 m/s$.
We carry out 2 trials of each level of the navigation task to test the baselines and average the results to evaluate the performance. 
% To more directly measure the discrepancy between simulated and real trajectories, 
% we use Wasserstein distance to calculate the distribution difference between simulated and real data as the trajectory distribution difference TrajGap.
% Two scenarios are selected for each level of navigation task to be tested, and the results in the table are the averages for both scenarios. 
From the results in the table, it can be seen that the transfer of the baselines from the simulation to the real environment has little influence on SR 
but has a significant impact on the mobile navigation process of the robot. 
Domain randomization methods can significantly improve the SPL and SFPL performance of the algorithm in the real world. 
Among them, UDR, as a simple randomization method, has smaller SPL and SFPL gaps in all three scenarios. 
The adaptive domain randomization methods have a significant advantage in SFPL performance in the real evaluation environment and have a smaller TrajGap.
GAT can effectively reduce the difference between the simulation velocity and the real velocity but does not necessarily improve the SPL and SFPL properties, 
probably because the algorithm utilizes the fitting error of the action transformer, which leads to poor generalization performance.
{
   SPL gaps and TrajGaps of the policies trained with None and GAT increase with the task level, while there is no significant trend in SFPL. 
   SPL and SFPL gaps of domain randomization have no significant trend with the various task levels. 
}

\subsubsection{Effect of maximum speeds on sim2real transfer performance (\textbf{Q4})}
Considering that the trajectory of the robot is affected by the speed, 
we analyze the influence of the maximum speed of the robot on the policy transfer from simulation to {the} real environment. 
We carry out 2 trials at Level 1 for each method and each scenario, where the robot starts to pose and the goals of the simulation are the same as the real environment.

From the results in the simulation environment, while SPL and SFPL of UDR are not significantly affected by the increase in speed. 
% For the SimOpt method, the increase in speed has no significant impact on the performance of SPL, but has an impact on the performance of SFPL and decreases with the larger speed. 
the change of speed has a significant impact on SPL and SFPL of other methods, and both decrease with the larger speed. 
From the results in the real environment, SPL and SFPL of almost all methods decrease with increasing speed. 
TrajGap also increases with the increase of velocity.

For a clearer view of the relationship between the sim2real gap and the maximum velocity, 
we have drawn the simulation and real performance differences of different metrics of different methods over different velocities in Fig. \ref{fig_speed_results}. 
From the results in the figure, SPL and SFPL gaps of SimOpt and GAT increase rapidly with velocity, 
especially from $1.0 m/s$ to $1.5 m/s$. 
SPLs of UDR and DROID have a severe decrease with increasing speed in the simulation environment, which leads to the opposite trend of their SPL gaps from SimOpt and GAT.
The SPL gap of the non-sim2real method also increases with velocity. 
For all methods, the trajectory discrepancy between the simulation and the real increases with increasing velocity. 
From the above results, we conclude that 
{
   with the increasing velocity, the performance of algorithms in simulation and real environment decreases, 
   and TrajGaps of all methods increase. 
}

\begin{figure*}
   \hspace{0.5cm}
	\centering  
   \includegraphics[width=0.95\textwidth]{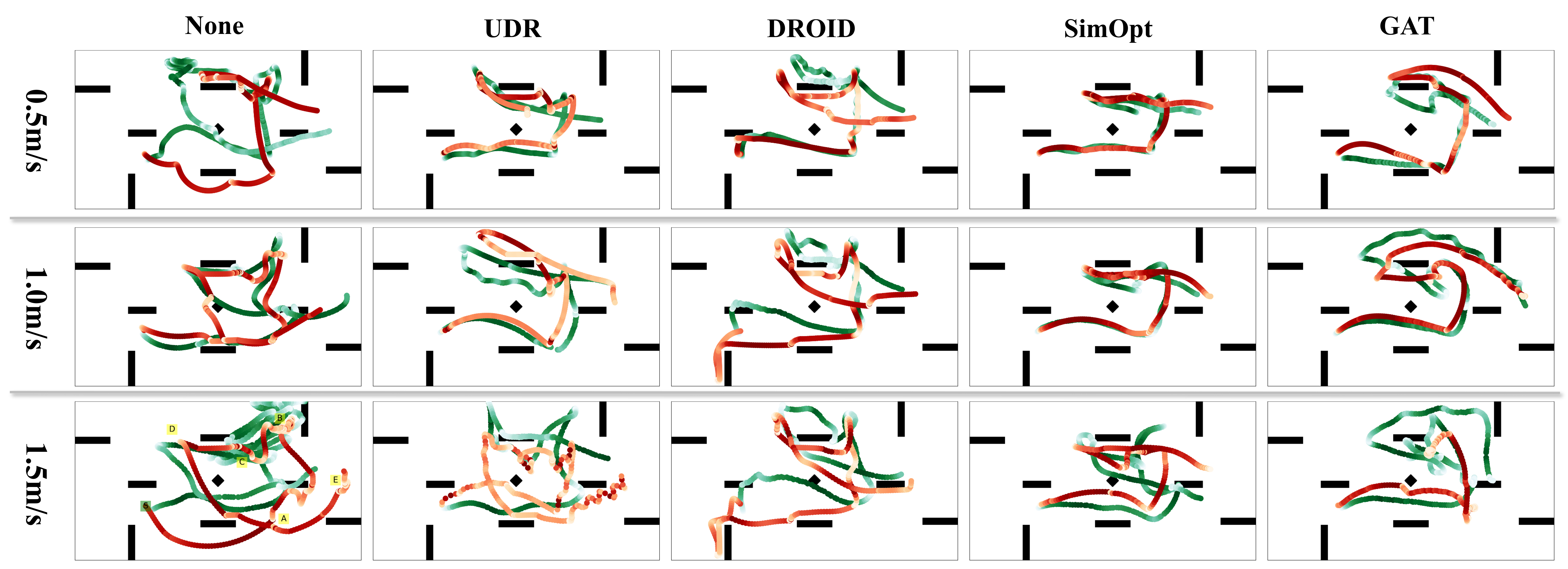} 
   
	\caption{Visualization of trajectory distribution with various maximum speeds.
   {
   The yellow squares are the goal blocks, 
   and the green square is the starting point of the agent. The red and green trajectories are collected in the simulation environment and the real environment, respectively.
   The color shade indicates the velocity, the darker the color, the larger the velocity. }
   }
   \label{fig_speed_trajectories}
   \vspace{-0.5cm}
\end{figure*}

In order to analyze the sim2real gap of the baselines at different velocities more clearly from the view of data distribution, 
we visualize the trajectories of the baselines, and the results are shown in Fig. \ref{fig_speed_trajectories}. 
{
From the results in the figure, it can be seen that the trajectories of the policy without the sim2real method have a clear distinction. 
% especially in the high-dimensional State-Action (S-A) space, the simulated and real data can be easily distinguished. 
For the adaptive domain randomization method and action transformation method, the trajectory distribution of the simulation is similar to the real. 
% and it is difficult to distinguish between the two types of data in high-dimensional space. 
It is noteworthy that as the velocity increases, it can be seen that the discrepancy between the simulated trajectory and the real trajectory becomes larger. 
% The data aggregation becomes stronger in high-dimensional space, and the two-state distributions gradually become distinguishable.
}
There are many reasons why the difference in trajectory distribution becomes larger with increasing velocity. 
First, when the maximum speed of the robot increases, the rotational acceleration of the motor output also increases, and slippage is more likely to occur. 
It is difficult for the simulation model to simulate the phenomenon. 
Second, when the desired velocity {increases}, the load on the motor increases, especially when the robot collides or slides with an obstacle at a higher speed, 
and the motor characteristic curve jumps out of the linear interval, which makes the simulation model fail. 
In addition, an increase in speed amplifies the effect caused by the control delay, which is a random variable, thus increasing the randomness of the trajectory. 
Finally, increasing speed will make the motion distortion of the point cloud more significant, 
and the difference between the corrected and simulated LiDAR data will become larger with increasing speed, 
and the difference of the input states will become discrepant. 
This state distribution shift makes the trajectory distribution different.

\subsection{{Time Efficiency of the Simulator}} 
{
We have collected the time cost of the simulator under different parallel numbers and different render modes. The experiments are conducted on an Intel i9-13900K CPU and an NVIDIA RTX 3070 GPU. The Ray framework is employed to implement parallel testing. During the testing, each thread completes an episode (1000 steps). This means that when the number of parallel environments is 8, we will collect 8000 tuples after running the test. The results are shown in Table \ref{table_time}. It can be seen that with the headless mode, it only takes 5 seconds to complete the whole episode. With render mode which renders images during the simulator running, the time cost will increase to 11.04 seconds. When the number of parallel environments is less than 8, the number of parallel environments does not significantly increase the time cost.
}

\begin{table}
   % \vspace{-0.5cm}
   \hspace{0cm}
   \center
   \caption{{Time efficiency of the simulator under different parallel numbers(N) and rendering modes. The time unit is second.}}
   \begin{tabular}{c c c c c c}
      \hline
        {Modes}   & {N=1} & {N=4} & {N=8} & {N=16} & {N=32} \\
      \hline
      {Headless mode}& {5.12} & {5.51} & {6.53} & {10.98} & {19.25}\\
      {Render mode}  & {11.04} & {13.13} & {13.88} & {16.67} & {26.80}\\
      
      \hline
   \end{tabular}
   \label{table_time}
   \vspace{-0.5cm}
\end{table}

\section{Conclusions and Future Application }
In this paper, we construct a hybrid framework, including the simulation system and the physical platform, for training and evaluating robot navigation policies. 
In the simulator, we provide a variety of sensors and rich dynamic interfaces for policy transfer. 
We also propose a novel safety metric named SFPL for evaluating the safety of the mobile robot. 
By evaluating common sim2real methods in this framework, we provide a new benchmark for robot navigation tasks. 
In addition, relying on this platform, we hold the 2022 IEEE CoG RoboMaster sim2real challenge to promote the development of flexible and agile agents.
We believe that there are many areas that will benefit from our proposed platform.
\begin{itemize}
   \item \textbf{Sim2Real Transfer}: The simulator of NeuronsGym provides a rich {dynamic} interface, including friction parameters, motor characteristics parameters, etc., 
   and a variety of sensor models to support sim2real algorithm development. {In addition, users can modify and design the layout of the simulation environment according to their needs and physical scenarios, in order to train agents suitable for their situation.} Researchers can study the generalizability of algorithms through sim2sim by setting different parameter distributions
   or use the physical platform to study sim2real problems. {In the future version, we will add a random layout generation plugin for training agents with the generalization of layouts.}
   \item \textbf{Safety Navigation Learning}: As mentioned in our paper, safety is vital for physical systems, and how to learn and evaluate safe policies is critical for real robot learning. 
   In mobile robotic systems, collision is the most common safety issue. We implement collision detection in both the simulated and real environments of NeuronsGym 
   and propose a more reasonable safety evaluation metric that can help researchers develop safe navigation algorithms more easily.
   \item \textbf{Visual Navigation}: Visual navigation is one of the {most} hotly developed fields nowadays. Although visual navigation is not introduced in this paper, 
   we also provide first-person view images in our platform, as well as the arena models with higher visual fidelity.
   We also organize the vision-based track in the 2022 IEEE CoG Sim2Real Challenge, but very few teams are able to complete it successfully. 
   Vision-based navigation is still a challenging task.
    \item {
    \textbf{Competitive Multi-Agent Policy Learning}: Unlike most robotics platforms, our NeuronsGym can support robot confrontation tasks. Competitive multi-agent reinforcement learning algorithms have made landmark advances in virtual games, but are difficult to feed back into physical systems. 
    We hope our robot confrontation tasks make physical robot policy learning benefit from the development of gaming algorithms in the virtual world.}
    \item {
    \textbf{Multi-Task Learning}: While multiple tasks in most current robot grasping platforms refers to grasping different objects, 
    the multiple tasks defined in our platform differ in task form, reward function, and valid action space. 
    This relaxed task form poses a greater challenge to multi-task learning algorithms, and we hope that this platform will facilitate the development of multi-task reinforcement learning algorithms.}
\end{itemize}

Of course, {we also note that fixed environment layouts make it difficult for the agent to cope with the challenges posed by real-world environmental changes.} 
In future work, we will explore larger arenas and more diverse environment layouts, as well as consider the inclusion of dynamic participants that can interact in the environment. 
In addition, we will also expand to {include} more agents to support research on multi-robot collaborative policy learning as well as intelligent emergence.

{
\bibliographystyle{IEEEtran}
\bibliography{ref}
}

\end{document}